\documentclass{article}

    \PassOptionsToPackage{numbers, compress}{natbib}

    \usepackage[final]{neurips_2024}

\usepackage{natbib}
\usepackage[utf8]{inputenc} 
\usepackage[T1]{fontenc}    
\usepackage{hyperref}       
\usepackage{url}            
\usepackage{booktabs}       
\usepackage{amsfonts}       
\usepackage{nicefrac}       
\usepackage{microtype}      
\usepackage{xcolor}         
\usepackage{graphicx}
\usepackage{subfigure}
\usepackage{booktabs} 
\usepackage{amsmath}
\usepackage{pifont}
\usepackage{xspace}
\usepackage{enumitem}
\usepackage{appendix}
\newcommand{\name}{VATT\xspace}

\title{Tell What You Hear From What You See - \\ Video to Audio Generation Through Text}

\author{Xiulong Liu \thanks{Department of Electrical \& Computer Engineering, University of Washington, Seattle, USA.} \\
\And
Kun Su \footnotemark[1] \\
\And
Eli Shlizerman \thanks{Department of Applied Mathematics, University of Washington, Seattle, USA}  
\footnotemark[1] \thanks{Corresponding author: shlizee@uw.edu}
}

\begin{document}

\maketitle

\begin{abstract}
The content of visual and audio scenes is multi-faceted such that a video stream can be paired with various audio streams and vice-versa. Thereby, in video-to-audio generation task, it is imperative to introduce steering approaches for controlling the generated audio. While Video-to-Audio generation is a well-established generative task, existing methods lack such controllability. In this work, we propose \textit{\name}, a multi-modal generative framework that takes a video and an optional text prompt as input, and generates audio and optional textual description (caption) of the audio. Such a framework has two unique advantages: i) Video-to-Audio generation process can be refined and controlled via text which complements the context of the visual information, and ii) The model can suggest what audio to generate for the video by generating audio captions. \name consists of two key modules:  \textit{\name Converter}, which is an LLM that has been fine-tuned for instructions and includes a projection layer that maps video features to the LLM vector space, and \textit{\name Audio}, a bi-directional transformer that generates audio tokens from visual frames and from optional text prompt using iterative parallel decoding. The audio tokens and the text prompt are used by a pretrained neural codec to convert them into a waveform. Our experiments show that when \name is compared to existing video-to-audio generation methods in objective metrics, such as VGGSound audio-visual dataset, it achieves competitive performance when the audio caption is not provided. When the audio caption is provided as a prompt, \name achieves even more refined performance  (with lowest KLD score of 1.41). Furthermore, subjective studies asking participants to choose the most compatible generated audio for a given silent video, show that \name Audio has been chosen on average as a preferred generated audio than the audio generated by existing methods. \name enables controllable video-to-audio generation through text as well as suggesting text prompts for videos through audio captions, unlocking novel applications such as text-guided video-to-audio generation and video-to-audio captioning.

\end{abstract}

\begin{figure}[tp]
    \centering
    \includegraphics[width=0.8\linewidth]{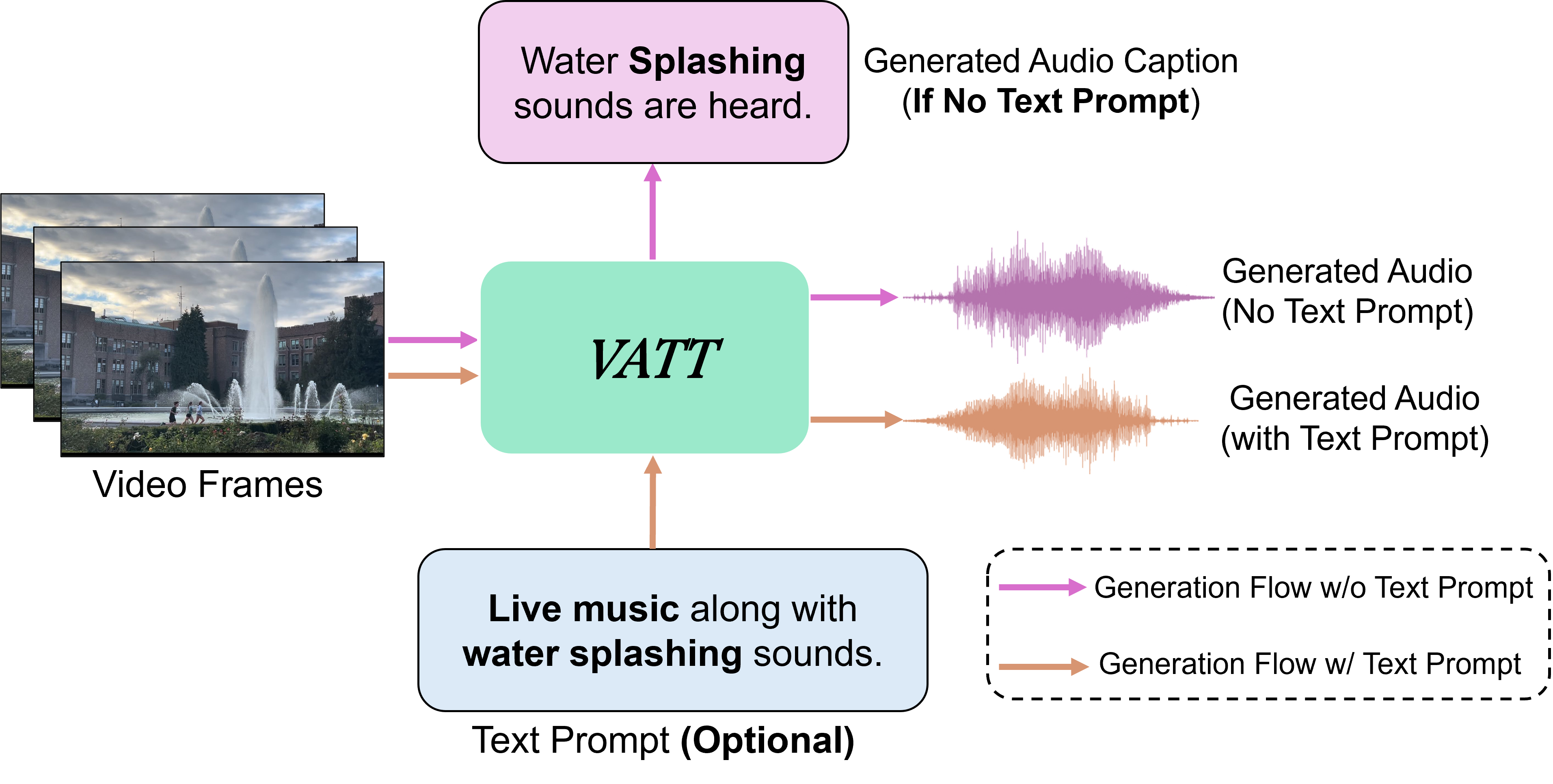}
    \caption{ \name is a flexible audio generative model capable of generating audio in two modes: i) When a silent video is the sole input, the model generates the audio along with a caption describing the possible audio that could match the video. ii) When in addition to the video, a text prompt is provided, the model generates audio aligned with both the video and the given text prompt.}
    \vspace{-3mm}
    \label{fig:teaser}
\end{figure}

\section{Introduction}
The combination of human perception and cognition represents a ``multi-modal'' way of processing and interpreting scenes. For example, when we are presented with a silent video of a fountain show attended by a crowd of people gathered around the spectacle our interpretation might translate the visual scene into an auditory experience, where the visuals are semantically processed and transformed into a corresponding sound narrative in our mind. Thus, we may associate audio that mixes sounds of splashing water accompanied by people talking and laughing with possibly background music in sync with the fountain.

As generative AI continues to progress, the incorporation of the aforementioned aspects into generative platforms presents itself as the future desirable capability. In particular, the goal of an ideal video-to-audio generative model would be to generate sounds that seamlessly match the video temporally and fully capture the semantics. Moreover, it is desirable to control such a generation process towards the themes and sounds that match user preference. Recent state-of-the-art approaches have adopted two types of generative modeling techniques: auto-regressive token-based modeling and diffusion-based modeling. These methods enable end-to-end video-to-audio generation and are applicable across a wide variety of video and audio categories. However, while these methods are capable of capturing the general semantics of sound sources in videos, they often overlook the subtleties of the context. For example, in a video depicting two cats in a territorial dispute, the model might produce a calm, amiable meowing sound, which contradicts the contentious nature of the scene. This discrepancy mainly stems from the limitations of the vision encoder, which struggles to distinguish between varying sound properties emitted by identical sound sources across different contexts, due to an incomplete understanding of the entire scene. Second, these methods lack controllability since the generation is conditioned only on visual frames, without taking into account the context and interpretation of the sounds. While text-to-audio models could explicitly control the context of the sounds, such models are based on text only without incorporating the rich and dynamic context of visuals, which could significantly inform video and audio alignment. Indeed, text only generative outcomes often result in unmatched audio with the visual (e.g., temporal misalignment or semantic loss).

To solve the above challenges, we propose a novel framework, Video-to-Audio Through Text (\name), that is able to generate audio from both video frames and an optional text prompt describing the expected sounds. \name consists of two modeling stages: i) Video-to-Caption stage,  which converts video features into an audio caption through a pretrained large language model (LLM) with a learnable projection layer. Through this cross-modal conversion, visual features that are relevant to audio concepts are extracted. These features are closely connected to audio-related tasks such as audio captioning and audio generation.  ii) Video + Text to Audio stage, that generates audio conditioned on the hidden states extracted from the LLM in the prior modeling stage. At its core, the proposed model in this stage is a bi-directional transformer decoder that generates audio using a token-based representation similar to~\cite{chang2022maskgit, borsos2023soundstorm}. To obtain the conditioning on the hidden states of the preceding component, the projected video features along with the optional text prompts are concatenated together and fed into the LLM in stage i), with the hidden states from the last layer extracted and attached to the audio tokens for the decoder. The decoder is trained using masked token modeling, where the objective is to predict masked audio tokens from unmasked ones at varying masking ratios. During inference, starting from all tokens being masked, an efficient parallel decoding algorithm is implemented which gradually unmasks multiple tokens in parallel based on video and text inputs until a stop condition is met. Finally, the generated tokens are converted into audio waveforms through a neural audio codec decoder.

We perform experiments with the proposed framework on existing large-scale audio-visual datasets such as VGGSound~\cite{chen2020vggsound} and Audioset-2M~\cite{gemmeke2017audio}. To facilitate training and evaluation with text, we created ``V2A Instruction'', a large-scale synthetic audio captions corpus, by prompting LTU-13B, an existing Audio LLM~\cite{gong2023listen}, to generate audio descriptions for both datasets. Our experiments demonstrate that the proposed model and its training method achieve competitive performance in comparison to previous video-to-audio methods on both objective and subjective metrics. Furthermore, it is designed to enable a controllable generation that adheres to both the video inputs and the text prompts. Indeed when a text prompt is provided, our experiments show significant improvement in audio metrics that measure the match of the generated sounds to the video. In addition, when the text prompt is not provided, our method can generate reasonable audio captions, which can be utilized for a potential description of the video or classification of sounds for a given video. These capabilities hence make \name a multifaceted single model able to perform both text-guided video-to-audio generation and video-to-audio captioning. 
To summarize our contributions:
\begin{itemize}
[align=right,itemindent=0em,labelsep=2pt,labelwidth=1em,leftmargin=*,itemsep=0em]
    \item To our best knowledge, we propose a first-of-its-kind framework that enables both text-guided video-to-audio generation and video-to-audio captioning through the integration of LLM.
    \item We create a large-scale synthetic audio captions dataset that facilitates text-conditional training and generation.
    \item Our method achieves state-of-the-art video-to-audio generation performance when compared with existing methods and enables text-controllable generation. In particular, our text-guided model surpasses existing SOTA in terms of KLD score (with lowest KLD score of 1.41) by a significant margin.
    \item \name generates audio in an efficient way - an order of magnitude faster than existing methods.
\end{itemize}

\section{Related Works}

\subsection{Visual-to-Audio Generation}
Visual-to-Audio Generation task has drawn significant attention since generative frameworks such as diffusion and transformer-based architectures have been developed. Existing Visual-to-Audio generation approaches can be divided into two branches of studies based on audio categories: \textit{visual-to-music} generation and \textit{visual-to-natural} sound generation. In visual-to-music generation domain, earlier studies explored Midi or spectrogram generation from human body movements by studying the temporal and semantics alignment~\cite{su2020audeo, gan2020foley, su2020multi, su2021does, Liu_2024_WACV}. More recently, diffusion-based methods have been proposed to generate music waveforms directly from videos~\cite{zhu2022discrete}. In visual-to-natural sound generation, earlier efforts pioneered the generation of sounds linked to various objects and materials~\cite{Owens_2016_CVPR}. Later works proposed an audio generation approach based on SampleRNN~\cite{zhou2018visual, mehri2016samplernn} that could generate several types of natural sounds from in-the-wild videos. While these approaches showcase promising results, they are often limited to specific audio categories. Neural codec~\cite{NIPS2017_7a98af17, esser2021taming, defossez2022high, kumar2024high} and autoregressive transformer architectures~\cite{NIPS2017_3f5ee243, radford2019language} addressed these limitations and as they have evolved, generative models now effectively generalize across a broader range of sounds or music, leveraging compressed latent spaces~\cite{iashin2021taming, sheffer2023hear, mei2023foleygen, su2023v2meow}. Similar advances have been shown with diffusion techniques such as~\cite{luo2024diff, wang2024v2a}. However, these methods often lack detailed sound control and their inference time turns out to be consuming. Our work aims to address these limitations by introducing a text-guided framework to improve controllability and efficiency in video-to-audio generation. While there are several concurrent works that aim to achieve partially similar goals to our proposed method~\cite{su2024vision,mo2023diffava,ziv2024masked}, our work is different since it is designed to achieve these capabilities within a single unified framework.

\subsection{Text-to-Audio Generation} As an alternative to the generation of audio from video, text can be used as an input to guide audio generation. When text is the input, audio generation becomes more controllable semantically. Existing approaches such as Make-An-Audio~\cite{huang2023make},  AudioLDM~\cite{liu2023audioldm}, AudioLDM-2~\cite{liu2023audioldm2} and others~\cite{dong2023clipsonic, huang2023noise2music,huang2023noise2music,schneider2023mo} enable general text-to-audio (or music) generation by adapting latent diffusion techniques, first developed in~\cite{rombach2022high}. Concurrently, methods such as AudioGen~\cite{kreuk2022audiogen}, MusicGen~\cite{copet2024simple}, AudioLM~\cite{borsos2023audiolm}, MusicLM~\cite{agostinelli2023musiclm},  SoundStorm~\cite{borsos2023soundstorm}, VampNet~\cite{garcia2023vampnet} leverage transformer-based architectures and token-based modeling techniques to produce audio tokens, that are then decoded into waveforms using neural codecs like Encodec~\cite{kumar2024high} and SoundStream~\cite{zeghidour2021soundstream}. Notably, SoundStorm and VampNet use an efficient technique known as masked token-based modeling which speeds up generation with parallel unmasking in the decoder. In our work, we consider a similar approach. While these models deliver high-quality audio with strong relevance to the text, they do not necessarily align with visual dynamics when adapted to video-to-audio generation. This is expected since such models have not been trained to attend to visual inputs. Our work addresses this by integrating a pretrained large language model (LLM) as a multi-modal encoder that processes both visual and textual inputs such that the generated audio considers both visual and text information.

\subsection{Multi-modal Large Language Models} Multi-modal Large Language Models (MLLMs), have been able to attain significant progress. With the advent of open source, pretrained and instruction-tuned LLMs such as LLama~\cite{touvron2023llama}, Alpaca~\cite{alpaca}, Vicuna~\cite{vicuna2023}. In particular, when extending these LLMs into MLLMs, a pretrained modality-specific encoder extracts the features and then a projection layer maps these features into vectors of the same dimension as text embeddings of the corresponding LLM. This approach led to developments in visual LLMs~\cite{liu2024visual,zhang2023video}, audio LLMs~\cite{gong2023listen, NEURIPS2023_3a2e5889}, audio-visual LLMs~\cite{chowdhury2024meerkat} and showed improvement in multi-modal understanding tasks such as captioning~\cite{Xu2016MSRVTTAL} and question-answering~\cite{li2022learning,liu2024tackling}. Recent efforts have also focused on tasks such as multi-modal retrieval~\cite{Dong_2024_CVPR}, multi-modal embodied navigation~\cite{liu2024caven, zhou2024navgpt}, leveraging LLM's strong reasoning capabilities to interpret or improve the results. In terms of generation, several works~\cite{tang2024any, wu2023next} aimed at achieving any-to-any modality generation using LLMs as a central medium. While these methods have been successful in general modality-to-modality generation, they do not achieve particular end-to-end video-to-audio generation, with or without text guidance, which is the unique direction our work focuses on.

\section{Methods}
\begin{figure}[tp]
    \centering
    \includegraphics[width=1.0\linewidth]{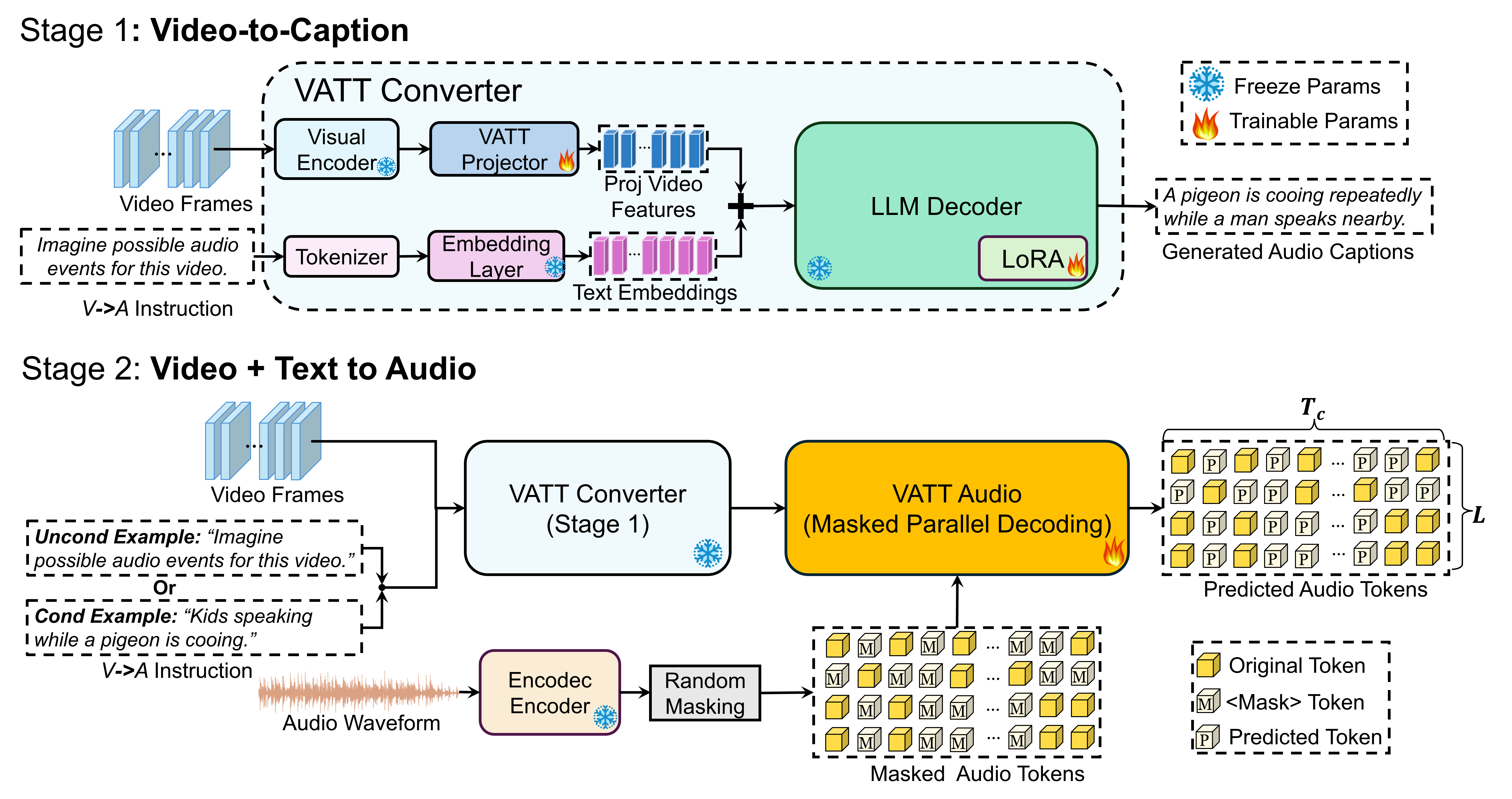}
    \caption{Two stages of \textit{\name} system training pipeline: (1) \textbf{Video-to-Caption} stage that maps video features into an audio caption through LLM. (2) \textbf{Video + Text to Audio} stage that learns to generate audio tokens through masked tokens prediction conditioned on Stage (1) features.}
    \label{fig:vatt_train}
    \vspace{-4mm}
\end{figure}

\name is a flexible vision-to-audio generative framework that can process both visual and textual inputs and generate both audio waveforms and captions of audio. To achieve this, \name consists of two modeling stages: i) \textbf{Video-to-Caption} : This stage utilizes a Large Language Model (LLM) with a learnable projection layer that converts video features into embeddings compatible with the LLM. The model receives an instruction to generate audio captions from video inputs. ii) \textbf{Video + Text to Audio}: This stage incorporates an encoder-decoder architecture. The encoder uses the finetuned LLM from Video-to-Caption stage with frozen weights. The decoder is a bi-directional transformer trained to generate audio tokens using masked token modeling techniques in training. The training pipeline of \name system is shown in Figure~\ref{fig:vatt_train}. During inference, \name generates audio tokens from video and optional text prompts through iterative parallel decoding. These tokens are then converted into audio waveforms using Encodec~\cite{defossez2022high}.

\subsection{Video-to-Caption Stage}

\textbf{\name Converter} is designed to integrate visual and textual prompts for audio generation as well as audio captioning. The core component, \textit{\name Projector}, is an embedding layer that maps video features into the text embedding space of the LLM. Given visual features extracted from frame-level vision encoders $V_f = \{v_1, v_2, ..., v_T\}$, a Linear layer is applied to project each feature from its original dimension $d_v$ to the LLM text embedding dimension $d_{lm}$, producing a sequence of transformed features $V_{lm} = V_f W_{l} + b_{l}$, where $W_{l}$ and $b_{l}$ are learnable parameters of the linear projection.

\textbf{V2A Instruction Tuning}: The key functionality of \name Converter is to extract from visual stream semantic features relevant to audio. Drawn on the success of multi-modal LLMs, such as visual-LLM~\cite{liu2024visual} and audio-LLM~\cite{gong2023listen}, we employ multi-modal instruction tuning to align the visual inputs of videos with the ground truth audio captioning of the same videos. Given a prompt instruction, $T_{i} = \{t_{i1}, t_{i2}, ..., t_{iK}\}$, such as ``Describe the audio that the video could generate:" and the projected visual features $V_{lm}$ as inputs, we model conditional distribution of audio descriptions $T_{a}=\{t_{a1}, t_{a2}, ..., t_{aN}\}$,  as $P_{\theta}(T_{a} | T_{i}, V_{lm})$ by fine-tuning an instruction-tuned LLM, e.g., Vicuna-7B~\cite{vicuna2023}. Unlike typical instruction-tuning that maps a signal into textual concepts within the same modality, our method bridges the concepts from visual to audio modality, unifying the representation for text-guided video-to-audio generation task that we describe in section \ref{sec:video_to_audio}. For training efficiency, we fine tune the LLM with \name Projector by integrating LoRA~\cite{hu2021lora} adaptors while keeping the original LLM weights frozen. We minimize the negative log-likelihood of audio caption tokens conditioned on visual inputs and prompt instruction
\begin{equation}
\mathcal{L}_{v2t}\left(\widehat{T_a} \mid T_i, V_{lm} \right)=-\sum_{l=1}^{N} log\left[P_\theta\left(\hat{t}_{al} = t_{al} \mid  T_{i}, V_{lm}\right)\right],
  \label{eq:generation_loss}
\end{equation}
where $t_{al}$ is the $l$-th text token in the ground truth audio description $T_{a}$, and $\theta$ is the set of trainable weights including \name Projector and LoRA adaptor. Further details of the constructions of text prompts and synthesis of audio captions are described in Section \ref{sec:dataset} and Appendix \ref{sec:v2a_instruction_dataset}. 

\subsection{Video + Text to Audio Stage}
\label{sec:video_to_audio}

Once the audio-related visual features are aligned with the text features in the LLM embedding space, the LLM effectively encodes multi-modal information that serves as a representation for text generation and audio generation. Indeed, in the second stage of \name, there are two generation modes to generate audio: i) When no conditional text prompt is provided, the video features along with a \textit{standard template} prompt (e.g., ``Describe possible audio that the video could infer.'') are fed as inputs to \name Converter. ii) When an audio caption is provided as the text prompt, the video features and the audio caption are fed together into \name Converter. In such a case, the provided audio caption helps guide the video-to-audio generation process and overrides the need for generated audio caption.  

\subsubsection{Audio Token Decoder}

To generate audio, we design an audio token-based decoder, \name Audio,  conditioned on the encoded features from \name Converter. In contrast to existing methods, which typically use auto-regressive token modeling~\cite{kreuk2022audiogen, copet2024simple, mei2023foleygen} or latent diffusion techniques~\cite{liu2023audioldm, liu2023audioldm2}, we adopt a novel token-based modeling technique based on masking tokens. The method, originally derived in image generation tasks~\cite{chang2022maskgit} and recently adapted to text-to-audio generation~\cite{borsos2023soundstorm,garcia2023vampnet}, is capable of achieving competitive generation quality while improving efficiency through an iterative parallel decoding algorithm during inference.

\textbf{Token-based Representation for Audio}
To represent audio waveforms using discrete tokens, we adopt a pretrained audio neural codec, Encodec~\cite{defossez2022high}, similarly to FoleyGen~\cite{mei2023foleygen}. Encodec is a multi-level residual vector-quantized (RVQ) autoencoder trained with waveform reconstruction and adversarial objectives, capable of high-fidelity reconstruction from compressed tokens. Specifically, Encodec uses $L=4$ codebooks of tokens to represent the audio. Lower-level codebooks encode coarse semantic information, while higher-level codebooks capture fine-grained details. We adopt an open source Encodec model pretrained using audio waveforms at $Sr_w = 16kHz$ sampling rate. The model compresses a waveform into tokens at $Sr_t=50Hz$ sampling rate, leading to  $r_{tw} = \frac{Sr_w}{Sr_t}  = 320$ waveform samples per token. For any waveform $A_{wav} \in \mathbb{R}^{1 \times T_w}$, we extract corresponding audio tokens representation ${A}_{tok} \in \mathbb{N}^{L \times T_c}$ ($ T_c = \frac{T_w}{r_{tw}}$) from Encodec encoder part. Once the model generates ${A}_{tok}$, the embedding vectors of $L$ levels of tokens at each time step are summed up before being sent to Encodec decoder to obtain the waveform.

\textbf{Masked Audio Token Generative Modeling} We model the distribution of audio token matrix $A_{tok} \in \mathbb{N}^{L \times T_c}$ by developing a token masking strategy which learns the joint distribution of the audio tokens in full parallelism. This is different than using ``delayed patterns'' proposed in~\cite{copet2024simple} which enables parallelism but only  on the level of codebook dimension. At each time step of $A_{tok}$, embedding vectors of $L$ tokens are summed up to represent audio waveform at the corresponding segment. In order to perform masking operation at any position, we introduce an additional learnable <MASK> token in each codebook. By randomly replacing some of the tokens entries in the $A_{tok}$ with <MASK> at corresponding codebook we obtain the masked audio token matrix $A^{M}_{tok} \in \mathbb{N}^{L \times T_c}$. We obtain $E^{M}_{a} \in \mathbb{R}^{d_{em} \times T_c}$ by summation of the embedding vectors of each token in $A^{M}_{tok}$ along the level axis.

\begin{figure}[tp]
    \centering
    \includegraphics[width=0.85\linewidth]{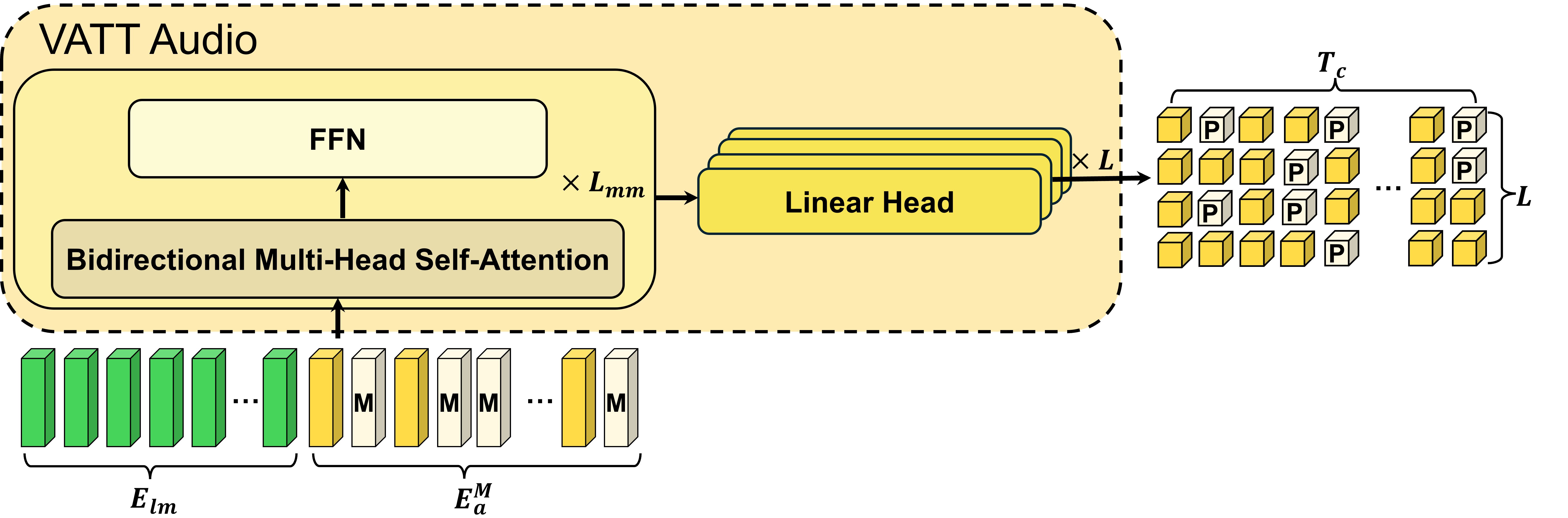}
    \caption{Audio Tokens Decoder: \name Audio is a bi-directional transformer that models the audio tokens and the conditioning inputs (LLM hidden states) jointly. We extract the part that corresponds to the audio features and apply $L$ Linear layers in parallel to perform classification on masked tokens at each codebook layer.}
    \label{fig:audio_decoder}
    \vspace{-3mm}
\end{figure}

Conditional generative modeling is implemented as follows. We extract the hidden states of the last layer $H_{lm} \in \mathbb{R}^{d_{lm} \times T_{lm} }$ (before the LLM prediction head) from \name Converter as the conditional inputs into the audio token decoder. We use a linear layer to project $H_{lm}$ to $E_{lm} \in \mathbb{R}^{d_{em} \times T_{lm}}$ with same feature dimension as the masked audio embeddings $E^{M}_{a}$. A straightforward way to model the relationship between $E^{M}_{a}$ and $E_{lm}$ is to use an interleaving self-attention and cross-attention block as proposed in Vanilla Transformer architecture~\cite{NIPS2017_3f5ee243}. However, we find that such interleaved interaction between audio and multi-modal input condition does not capture the fine-grained correspondence between them. Therefore, we propose to use a bi-directional self-attention architecture to fuse the features. 

Specifically, we concatenate $E_{lm}$ with $E^{M}_{a}$ along the temporal axis to obtain the fused features $E_{mm} = Concat(\left[E_{lm}, E^{M}_{a}\right])$. The decoder consists of $L_{mm}$ layers of self-attention blocks, as shown in Fig.~\ref{fig:audio_decoder}. The output hidden states in the last layer of the decoder, $ H^{out}_{mm} = Dec\left(E_{mm}\right)$, represent fused audio and conditions features. We only extract the part of the hidden states corresponding to the audio tokens, $H^{out}_{a} \in \mathbb{R}^{d_{mm} \times T_{c}}$, and pass it through $L$ Linear layers in parallel to perform classification on masked tokens at each level of the codebooks. For each masked audio token in matrix $A^{M}_{tok}$, we calculate the cross-entropy loss between the predicted token $\hat{a}_{tok}$ and the ground truth token $a^{gt}_{tok}$, formulated as 
\begin{equation}
\mathcal{L}_{\name}=-\sum\limits_{\substack{a_{tok} \in A^{M}_{tok}}} \mathbb{I}\left(a_{tok} = \text{<MASK>} \right) log\left[P_{\phi}(\mathbf{\hat{a}_{tok}} = a^{gt}_{tok} | A^{M}_{tok}; H_{lm}))\right],
\end{equation}

where $\phi$ is the set of trainable parameters in the audio token decoder, and $\mathbb{I}$ is the indicator function.

\subsubsection{Masking Design and Iterative Parallel Decoding} \textbf{Masking Distribution Design} Inspired by~\cite{chang2022maskgit, borsos2023soundstorm}, we incorporate variable random masking. In particular, it was shown that masking ratio plays an important role in audio token decoder to generate meaningful signals. 
While in~\cite{chang2022maskgit, borsos2023soundstorm} arc-cosine masking distributions is used by default, here we study several masking strategies that include distributions along with different hyper-parameters to find the strategy that reaches more optimal generation quality (see Appendix \ref{sec:additional_ablation} for further details). Our study shows that normal distribution with a mean of 0.75 and standard deviation of 0.25, truncated from 0.5 to 1.0 is such optimal strategy. The general interpretation of this strategy is that a relatively high range masking ratio enables models to generate better initial tokens when most of the entries in the token matrix are masked. This is essential for future decoding steps to generate meaningful tokens.

\textbf{Iterative Parallel Decoding} Scheduling of masking plays a key role as well. During inference, we follow the cosine scheduling scheme proposed in~\cite{chang2022maskgit} to gradually resolve the audio tokens. The iterative sampling procedure starts with all <MASK> in the audio token matrix. At a step $t$, the model takes the audio token matrix $A_{t-1}$ from the previous step along with the conditions as inputs and samples a new audio token matrix $\hat{A}_{t}$ in parallel with all tokens unmasked. Based on the confidence at each entry of $\hat{A}_{t}$ only tokens with top-k confidence are kept while the remaining entries are re-filled with <MASK>, resulting in $A_{t}$. The cosine scheduling scheme determines the ratio of re-masked tokens by $r_{t} = cos\left(\frac{\pi}{2} \cdot \frac{t}{T}\right)$. Notably, to resolve the confidence of each entry in the matrix, we adopt the \textit{``gumbel-top-$k$ trick''}~\cite{kool2019stochastic} with temperature that varies, i.e.,  $c_i = \frac{log(p_i)}{\tau} + G$, where $G \sim \text{Gumbel}(0, 1)$ and $p_i$ denotes the output probability of the sampled token at the entry $i$. This is equivalent to sampling k values from multinomial distribution from the softmax probabilities without replacement. The temperature $\tau$ controls the degree of stochasticity. We use $\tau = \tau_{0} \cdot (1 - \frac{t}{T})$ with linear decay during generation, where $\tau_0$ is the initial temperature. Similarly to~\cite{chang2022maskgit, borsos2023soundstorm}, our method achieves optimal quality and fast speed within a few decoding steps (typically 10 - 20).

\section{Experiments}
\textbf{Datasets:} \label{sec:dataset}
We use common benchmarks datasets VGGSound~\cite{chen2020vggsound} and AudioSet-2M~\cite{gemmeke2017audio} for training and evaluation. VGGSound is a large-scale audio-visual dataset sourced from YouTube, containing 192k videos from 309 audio-visual categories, with 177k / 15k train-test video splits. AudioSet-2M is a larger audio-visual database with around 2M YouTube videos, with only 1.6M available online. In Stage 1, we train \name Converter with both datasets and test on VGGSound only. In Stage 2, for fair comparison against existing video-to-audio generation methods, we train and evaluate on VGGSound dataset only.

To train \name with text, we synthesize a large-scale audio caption dataset, ``V2A Instruction'', using LTU~\cite{gong2023listen}, an existing audio LLM. We obtain audio captions by prompting the pretrained LTU-13B model with the inputs of audio waveform along with the instruction \textit{``\#\#\# Instruction: Close-ended question: Write an audio caption describing the sound. \#\#\# Response:''}. For AudioSet~\cite{gemmeke2017audio} and VGGSound~\cite{chen2020vggsound} we generate a single audio caption per each video for a total of 1.77M videos.

To ensure the quality of captions, we first manually verified the validity of LTU-generated captions prior to using them as synthetic ground-truth (GT) and then performed an experiment to further evaluate captions quality. In particular, we randomly selected 100 videos from VGGSound test set with stratified sampling according to video categories to conduct a human study. We used 1-5 point MOS (Mean-Opinion-Score) scale (the higher the better) to measure correctness of the captions. We provide pairs of videos and the corresponding captions to the raters, asking ``How accurately the provided caption reflects the sound events happening in the video? 1. Inaccurate and irrelevant. 2. Relevant but inaccurate with many mistakes. 3. Partially accurate but missing details and with mistakes. 4. Mostly accurate with some minor mistakes. 5. Accurate and complete." We used the MTurk platform to perform the evaluation and collected a total of 300 responses. The generated captions have a high MOS of mean 4.72 and std 0.37, providing an additional indication for the validity of the synthetic ground truth.

\textbf{Implementation Details:} For visual inputs, we use eva-CLIP~\cite{sun2023eva} image encoder to extract mean-pooled visual features from video frames at 5fps rate, which result in $50 \times 768$ visual sequence for a 10s video. To represent audio, we extract audio tokens from a pretrained Encodec-16kHz. For each 10s audio waveform, we represent it with $A_{tok} \in \mathbb{N}^{4 \times 500}$ token matrix.

For LLM, we explore two open-source models, Gemma-2B~\cite{team2024gemma} and LLama-2-7B~\cite{touvron2023llama}, using instruction-tuned checkpoints. The LLM hidden size of Gemma-2B is 2048 and 4096 for LLama-7B. For both LLMs, we train \name Converter using LoRA parameter-efficient fine-tuning technique while keeping the LLM weights frozen. We use rank $r = 16$ and $\alpha=32$ with 0.1 dropout rate for LoRA configuration.

\name Audio is a bi-directional transformer with 24 layers, each with hidden size 1024 with 16 attention heads. To differentiate the conditioning inputs and audio tokens, we add two learnable modality-specific embeddings with respect to the corresponding inputs(see further implementation details in Appendix\ref{sec:additional_implementation}).
 
\textbf{Evaluation Metrics:} To evaluate video-to-audio generation quality, we follow the method of~\cite{mei2023foleygen}, which proposed the metrics Kullback-Leibler-Divergence (KLD) with PassT~\cite{koutini2021efficient}, Fréchet Audio Distance (FAD)~\cite{kilgour2018fr} and Align Accuracy (Align Acc)~\cite{luo2024diff}. KLD measures how closely the generated audio matches the GT through pairwise comparison, reflecting how well the audio captures the concepts in the video. FAD evaluates the overall distribution, indicating the overall quality of the audio. Align Acc assesses the relevance and temporal alignment of the audio and the video. Additionally, we incorporate generation speed (time taken per waveform sample) to measure efficiency.
We also compute the CLAP score~\cite{wu2023large} to evaluate the adherence of generated audio to text prompts to compare our results with text-to-audio generation. Further details of these metrics are described in Appendix~\ref{sec:metrics}. 

For video-to-audio captioning, we use two types of metrics, natural language generation (NLG) metrics and audio-text relevance metric. NLG metrics evaluate the generated captions with respect to the ground truth audio captions using rule-based matching in terms of precision and recall. These metrics include BertScore~\cite{zhang2019bertscore}, BLEU-4~\cite{papineni-etal-2002-bleu}, ROUGE-L~\cite{lin-2004-rouge} and CIDEr~\cite{vedantam2015cider}. To assess the 
relevance of generated audio captions with the actual audio, we compute the CLAP-score~\cite{wu2023large} as cosine similarity between audio and text embeddings.   

\textbf{Quantitative Evaluation of Audio Generation:} We evaluate audio generation of \name models on the VGGSound test split. For each of the 15,446 video samples, we generate a 10-second audio waveform. We compare \name variants against existing video-to-audio generation methods as well as text-to-audio generation methods including AudioLDM-2~\cite{liu2023audioldm2} and AudioGen~\cite{kreuk2022audiogen} using different text prompts. The results on the metrics described above are summarized in Table \ref{tab:v2a_comparison} and Table \ref{tab:t2a_comparison}. \textbf{\name models} achieve best KLD score and Align Acc against other methods while maintaining competitive FAD (top 2). Notably, when guided by GT audio captions (\name-LLama-T and \name-Gemma-T; bottom) our models generate sounds that match the GT audio more accurately, as indicated by lowest KLD score of 1.41 and 1.66 for \name models with two LLM backbones, surpassing both video-to-audio and text-to-audio methods. In comparison to text-to-audio methods, \name models achieve competitive audio-text alignment in terms of CLAP score, demonstrating a strong capability to follow text prompts. Implementation details of these baselines are included in Appendix \ref{sec:detail_baselines}.

\begin{table*}[t]

\caption{Quantitative results against video-to-audio generation methods on VGGSound test set. `-T' refers to model with text prompts.}
\label{tab:v2a_comparison}
\vspace{1mm}
\centering
\resizebox{0.65\linewidth}{!}{
\begin{tabular}{lcccc}
\toprule
Methods & KLD $\downarrow$ & FAD $\downarrow$ & Align Acc $\uparrow$ & Speed (s) $\downarrow$\\
\midrule
SpecVQGAN~\cite{iashin2021taming} & 3.78 & 6.63 & 48.79 & 7.2 \\
IM2WAV~\cite{sheffer2023hear}      & 2.54 & 6.32 & 74.31 & 289.5 \\
Diff-Foley~\cite{luo2024diff} & 3.15  & 6.40 & 82.47 & 4.4 \\
FoleyGen~\cite{mei2023foleygen} & 2.89 & 2.59 & 73.83 & 6.9 \\
V2A-Mapper~\cite{wang2024v2a} & 2.78   &   \bf{0.99}   & 74.37  &  11.54   \\
\textbf{\name-LLama (Ours)} & 2.39 & 2.38 & 80.32 & \bf{1.1} \\
\textbf{\name-Gemma (Ours)} & \bf{2.25} & 2.35 & \bf{82.81} & \bf{0.65} \\
\midrule 
\textbf{\name-LLama-T (Ours)} & \bf{1.41} & 2.54 & 80.16 & \bf{1.2} \\
\textbf{\name-Gemma-T (Ours)} & \bf{1.66} & 2.98 & 81.48 & \bf{0.76} \\
\bottomrule
\end{tabular}
}
\end{table*}

\begin{table*}[t]

\caption{Quantitative results comparing \name with text-to-audio generation methods on VGGSound test set. `-T' refers to model with text prompts. CLAP score is calculated as the cosine similarity of generated audio with respect to the GT audio caption.}
\label{tab:t2a_comparison}
\vspace{1mm}
\centering
\resizebox{0.85\linewidth}{!}{
\begin{tabular}{lccccc}
\toprule
Methods & Text Prompt & KLD $\downarrow$ & FAD $\downarrow$ & Align Acc $\uparrow$ & CLAP Score $\uparrow$ \\
\midrule
AudioGen~\cite{kreuk2022audiogen} & LLAVA visual caption & 3.65 & 6.03  & 41.66  & - \\
AudioGen~\cite{kreuk2022audiogen} &  GT audio caption  & 2.19 & 3.17 & 48.96  & \bf{0.409} \\
AudioLDM-2~\cite{liu2023audioldm2} & LLAVA visual caption  & 3.54 & 3.62 & 53.49 & - \\
AudioLDM-2~\cite{liu2023audioldm2} & GT audio caption & 2.09  & \bf{2.46} & 51.84 & 0.326  \\

\textbf{\name-LLama-T (Ours)} & GT audio caption & \bf{1.41} & 2.54 & \bf{80.16} & 0.347  \\
\textbf{\name-Gemma-T (Ours)} & GT audio caption & \bf{1.66} & 2.98 & \bf{81.48} & 0.310\\
\bottomrule
\end{tabular}
}
\end{table*}

\textbf{Quantitative Evaluation of Video-to-Audio Captioning:} We evaluate video-to-audio captioning by prompting \name Converter to generate audio captions. We use the prompt ``Describe the possible audio for this video:'' to generate captions for all VGGSound test videos. For baselines, we prompt LLAVA-13B-v1.5 model in two zero-shot modes to generate visual and audio descriptions respectively. Since LLAVA can  take a single image as an input only, we select the middle frame of videos. We use ``Provide a concise, descriptive caption for the following image.'' as the visual prompt, and ``Describe the sounds that this scene could yield in a short sentence without reasoning'' as the audio prompt. We also compare against a video LLM baseline, Video-LLAMA-7B, to perform zero-shot video-to-audio captioning. Specifically, we directly input VGGSound videos into the VL branch of the Video-LLAMA model, and prompt it to generate audio captions using the instruction “User/ What sounds could match the video?” Since Video-LLAMA has not been pretrained on VGGSound dataset and LTU generated captions, we implement a similar structure of Video-LLAMA and train on our LTU-generated captioning data. We replaced the original BLIP-2 visual features used by Video-LLAMA with our eva02-CLIP-L visual features due to the expensive pre-processing time for all BLIP-2 features from videos in VGGSound and AudioSet. For the Video-QFormer component of Video-LLAMA, we keep it the same as Video-LLAMA, and we name this model as VATT-Qformer - LLama. Our evaluation is summarized in Table \ref{tab:captions_comparison}. \name models with LLMs outperform LLAVA-prompted and Video-LLAMA zero-shot results demonstrating a stronger capability to infer sounds from videos semantically. In particular, when measuring audio-text relevance, our model with LLama achieves an increase of \textbf{+5.0\%} in accuracy when compared with LLAVA visual caption baselines. For reference, the ground truth audio captions generated by LTU~\cite{gong2023listen} have an average CLAP score of 0.379.

\begin{table*}[!h]
  \caption{Comparison of video-to-audio captions on NLG evaluation metrics and text-audio relevance (CLAP Score).}
  \label{tab:captions_comparison}
  \vspace{1mm}
  \centering
  \resizebox{0.9\linewidth}{!}{
  \begin{tabular}{@{}llcccccr@{}}
    \toprule
    Methods  & BertScore (F1)  $\uparrow$ & BLEU-4 $\uparrow$ & ROUGE-L $\uparrow$ & CIDEr $\uparrow$ & CLAP Score $\uparrow$ \\
    \midrule
    LLAVA w/ Visual Prompt & 0.855 & 0.089 & 0.137 & 0.026 & 0.213 \\
    LLAVA w/ Audio Prompt & 0.870 & 0.123 & 0.155 & 0.095 & 0.182 \\
    Video-LLAMA w/ Audio Prompt & 0.861 & 0.091 & 0.117 & 0.021 & 0.204 \\
    \name Converter - Gemma (ours) & 0.900 & 0.345 & 0.337 & 0.926 & 0.229 \\
    \name-Qformer - LLama & 0.907 & 0.419 & 0.375 & 1.264 & 0.245 \\
    \name Converter - LLama (ours) & \bf{0.909} & \textbf{0.424} & \bf{0.384} & \bf{1.354} & \bf{0.263} \\
    \bottomrule
  \end{tabular}
  }
\end{table*}

\begin{table*}[tp]
\caption{Architecture Ablation Study.}
\label{tab:arch_ablate}
\vspace{1mm}
\centering
\resizebox{0.5\linewidth}{!}{
\begin{tabular}{lcccc}
\toprule
Methods & KLD $\downarrow$ & FAD $\downarrow$ & Align Acc $\uparrow$\\
\midrule
\name-V & 2.43 & 2.53 & 82.43  \\
\name-Cross-Attn & 2.76 & 3.63 & 76.85 \\
\textbf{\name-Gemma (Ours)} & \bf{2.25} & \bf{2.35} & \bf{82.81} \\

\bottomrule
\end{tabular}
}
\end{table*}

\begin{figure}[tp]
    \centering
    \includegraphics[width=1.0\linewidth]{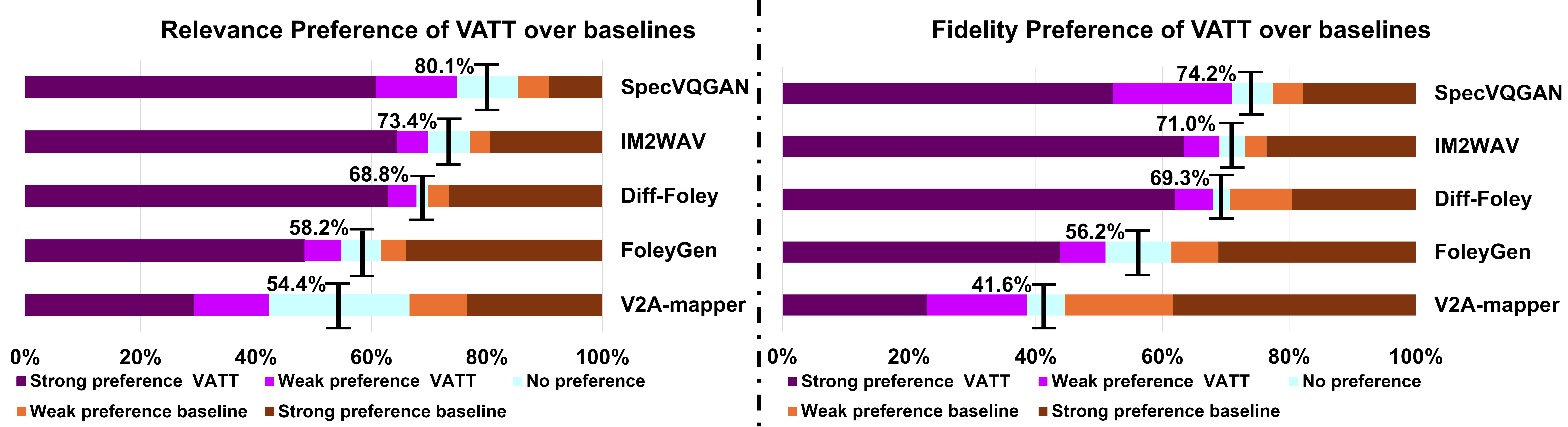}
    \vspace{-2mm}
    \caption{Qualitative evaluation results: Pairwise Comparison of generated audio \name v.s other methods comparing Fidelity and Relevance aspects.}
    \label{fig:human_eval}
    \vspace{-0.3cm}
\end{figure}

\textbf{Qualitative Evaluation:} In addition to quantitative evaluations, we also conduct a qualitative (subjective) study to evaluate audio generation perceptual quality of \name. Specifically, we randomly select 100 videos from VGGSound test split with stratified sampling according to video categories. For each method in the baseline, we pair the generated samples against \name. Two aspects of the generation are evaluated, Fidelity and Relevance. Fidelity focuses solely on audio quality, while Relevance evaluates the semantic relevance and temporal alignment of audio to the video. For each pair, raters are asked to rate their scoring of \name versus a compared baseline on a Likert scale from 1 (strongly prefer baseline) to 5 (strongly prefer \name). We use our best \name variant, \name-LLama-T (with GT text guidance), for the comparison. As shown in Table \ref{fig:human_eval}, \name surpasses other methods in Relevance. In terms of Fidelity, \name is consistently being preferred when compared with most baselines and slightly less preferred when compared with V2A-Mapper. The reason could be that V2A-Mapper is directly optimized with diffusion techniques on AudioLDM, a large-scale pretrained text-to-audio model, such models tend to perform better in fidelity aspect in comparison to token-based models. Further details of qualitative evaluations are incorporated in Appendix \ref{sec:human_eval_details}, and qualitative samples are provided in Appendix \ref{sec:qual_samples}.

\textbf{Ablation Studies:} 
We study the effectiveness of \name Converter by removing the LLM and directly feed the visual features into the decoder. We denote such model as \name-V. While \name-V does not handle textual inputs or generate text, it still serves as a strong variant of \name for video-to-audio generation.

To study the contribution of audio token decoder, we replace the decoder part of \name with interleaving attention blocks proposed in vanilla transformer~\cite{NIPS2017_3f5ee243}, and denote this variant as \name-Cross-Attn. 
As shown in Table \ref{tab:arch_ablate} \name-Gemma model outperforms both \name-V and \name-Cross-Attn. When \name is conditioned on visual inputs only its performance is lowest across variants.. The \name Converter enhances the visual features through audio-relevant text, thereby improving the relevance and quality of the generated audio. In addition, we find that the bi-directional transformer design in \name Audio is critical for learning the associations between audio and conditioning inputs to enhance audio generation performance. Additional ablation studies can be found in Appendix \ref{sec:additional_ablation}.

\section{Conclusion} In this work, we propose a multi-modal generative framework that enables both text-guided video-to-audio generation and video-to-audio captioning. Experiments show that our method can generate high quality audio through text in both unconditional and conditional modes, as well as to generate reasonable audio captions from videos. One area for improvement is the diversity of the text generated by current audio LLMs. In cases where the user-provided text prompts significantly differ in style there is a possibility for a conflict of audio quality and adherence to the instructions. Future work could enhance the capability of the model to generalize across different text styles and to further develop capabilities for informative iterative conversation-like video-to-audio generation.

\section*{Broader Impact}
\label{sec:broader_impacts}
\name could augment existing audio-video creation tools for content creators by allowing generation of custom audio tracks for given visual content through user provided text prompts. Also, \name has the ability to suggest potential sounds for a given video which can inspire creators by presenting audio options that may not have been considered otherwise. This feature can be useful for brainstorming of content creation, where audio choices can influence the style of the final product. 

Further extensions of this work could involve conversational video-to-audio generation such that the audio content is iteratively being refined. By integrating a conversational interface, the users can engage in a dialogue with the system, making requests and receiving responses. This approach goes beyond static text inputs, offering a more accessible toolset that does nto require significant audio editing expertise. Moreover, the conversational system can seek clarifications or propose alternatives, functioning like an assistant to avoid misunderstandings and enhance audio quality. More broadly, the generative approach proposed here has the potential to adapt to other generative areas not limited to audio, video, but also potentially impact fields such as biochemistry, physics where a generative approach is utilized, e.g., generative modeling of high-energy particle events~\cite{liu2024calo,krause2024calochallenge}.

While VATT presents a potential for content creation, the ability to generate realistic audio from visual inputs could lead to misuse, such as creating deceptive content or deepfake audio and ethical concerns must be addressed before utilization. Furthermore, similarly to audio generation, text generation capability could result in misuse such as offensive language or privacy violations. To mitigate these risks, in further development or potential code release we will establish clear ethical guidelines, evaluate for biases, and implement safeguards to ensure responsible use and fair outputs.

\begin{ack}
We acknowledge the support of HDR Institute: Accelerated
AI Algorithms for Data-Driven Discovery (A3D3) National
Science Foundation grant PHY-2117997 and the departments of Applied Mathematics and Electrical and Computer Engineering at the University of Washington.
\end{ack}

\clearpage
\newpage

{
\small
\bibliographystyle{neurips_2024}
\bibliography{neurips_2024}
}

\newpage
\appendix
\onecolumn

\section*{Appendix}

\noindent In this Appendix, we provide:
\begin{itemize}
[align=right,itemindent=0em,labelsep=4pt,labelwidth=1em,leftmargin=*,itemsep=0.5em]

    \item Additional Ablation Studies and Comparisons,  see Appendix \ref{sec:additional_ablation}.
    \item Qualitative Examples and Analysis, see Appendix \ref{sec:qual_samples}.
    \item Details and Examples of our synthetic ``V2A Instruction'' Dataset, see Appendix~\ref{sec:v2a_instruction_dataset}.
    \item Additional Implementation Details of \name, see Appendix \ref{sec:additional_implementation}.
    \item Additional Implementation Details of baselines, see Appendix \ref{sec:detail_baselines}.
    \item Details of Evaluation Metrics, see Appendix \ref{sec:metrics}
    \item Human Evaluation Details, see Appendix \ref{sec:human_eval_details}
\end{itemize}

\section{Additional Ablation Studies and Comparisons}
\label{sec:additional_ablation}
\textbf{Masking Ratio Distribution Ablation:} Designing an appropriate varying masking ratio distribution for training is essential to achieve high audio quality and relevance. We study several commonly used masking ratio distributions, including Uniform distribution, Gaussian distribution and arc cosine distribution. For Gaussian distribution, we experiment with distributions with 4 different mean values, 0.55, 0.75, 0.95, and moving mean following a sine schedule with respect to the training epoch (similar to curriculum learning), in the range of  $[0.25,0.95]$. The standard deviation is kept fixed at 0.25. We use \name Gemma-2B model for the masking ratio ablation study. As shown in Table \ref{tab:ablate_mask}, the model performs better when the distribution has a higher mean in masking ratio, especially arc cosine distribution and Gaussian mean with 0.75. This is due to the fact that the initial steps during the sampling stage are important for future decoding steps. The initial steps correspond to high masking ratio cases. For later steps, new tokens are unmasked conditioned on more clues such that the masking ratio decreases and the generation becomes less challenging, thus making the learning at lower masking ratio easier during training.

\textbf{Self-prompting Ablation:} When additional text prompts are provided as inputs, we could use the audio captions generated by the \name Converter as the text prompt to generate the audio. In this self-prompting mode, the generated audio could be interpreted by the same model in terms of the caption. As shown in Table \ref{tab:ablate_self_prompt}, when our model is fed with corresponding generated captions, the model performs slightly worse than the model without prompt input, showing the space for improvement in the quality of generated captions. Also, generation of audio with the captions from \name-Converter-LLama outperforms captions from \name-Converter-Gemma, in particular evident from the FAD and KLD scores.  As the caption quality improves, the text-conditioned video-to-audio generation performance also improves. The GT audio captions generated by LTU obtains the highest CLAP score (measured with respect to the GT audio in original video) of 0.379, reflecting the best caption quality. Feeding such GT captions as input to the model also leads to the best audio generation results.

\begin{table*}[p]
\footnotesize
\caption{Ablation on Masking Ratio Distribution.}
\label{tab:ablate_mask}
\vspace{1mm}
\centering
\begin{tabular}{lcccc}
\toprule
Methods & KLD $\downarrow$ & FAD $\downarrow$ & Align Acc $\uparrow$\\
\midrule
Uniform & 2.52 & 2.75 & 80.37  \\
Arc cosine & 2.34 & 2.26 & 82.88 \\
Gaussian w./ mean 0.55 & 2.31 & 2.34 & 82.27 \\
Gaussian w./ mean 0.75 & 2.25 & 2.36 & 82.81 \\
Gaussian w./ mean 0.95 & 2.24 & 2.49 & 81.80 \\
Gaussian w./ moving mean & 2.42 & 2.32 & 81.81 \\
\bottomrule
\end{tabular}

\end{table*}

\begin{table*}[tp]
\footnotesize
\caption{Ablation on self-prompting text-guided generation.}
\label{tab:ablate_self_prompt}
\vspace{1mm}
\centering
\begin{tabular}{lcccc}
\toprule
Methods & Text Prompt & KLD $\downarrow$ & FAD $\downarrow$ & Align Acc $\uparrow$ \\
\midrule

\textbf{\name-LLama} & \ding{55} & 2.39 & 2.38 & 80.32  \\
\textbf{\name-Gemma} & \ding{55} & 2.25 & 2.35 & 82.81  \\
\textbf{\name-LLama-T} & \name-Converter-LLama & 2.38 & 2.58 & 80.41  \\
\textbf{\name-LLama-T} & \name-Converter Gemma	& 2.57 & 2.67 &	79.20 \\
\textbf{\name-Gemma-T} & \name-Converter-Gemma & 2.40 & 3.67 & 80.07  \\
\textbf{\name-Gemma-T} & \name-Converter-LLama & 2.26 &	3.20 & 80.42 \\
\textbf{\name-LLama-T} & GT audio caption & 1.41 & 2.54 & 80.16 \\
\textbf{\name-Gemma-T} & GT audio caption & 1.66 & 2.98 & 81.48 \\
\bottomrule
\end{tabular}
\end{table*}

\textbf{Comparison with Text-to-Audio generation methods on AudioCaps:} We use VGGSound as our main dataset and benchmark to evaluate since it is a large-scale audio-visual dataset with around 200K videos across many categories, and also the quality of audio-visual alignment is high. To further show the generalization capability of \name, we experiment with AudioCaps dataset. Due to limited video samples in AudioCaps, we finetuned our VGGSound pretrained VATT model on AudioCaps dataset in two settings, with and without text prompts. To keep the comparison fair, we use the GT audio captions from AudioCaps as the text prompts. We use \name-LLama and \name-LLama-T to compare against AudioGen and AudioLDM-2. As shown in Table \ref{tab:audiocaps_result}, \name-LLama-T performs on a similar level to AudioGen in terms of FAD and KLD score, while falling behind AudioLDM-2. It is noteworthy that the audio decoder of both AudioGen and AudioLDM-2 are pretrained on much larger data scale (7000 hrs and 30000 hrs audio respectively) than ours (700 hrs audio). Despite this, \name still performs reasonably well on this dataset.

\begin{table*}[tp]
\caption{Quantitative results against text-to-audio generation methods on AudioCaps test set.}
\label{tab:audiocaps_result}
\vspace{1mm}
\centering
\small 
\begin{tabular}{lcccc}
\hline
Methods        & KLD ↓   & FAD ↓   & Align Acc ↑   & CLAP Score ↑ \\
\hline
AudioGen       & 2.09    & 3.13    & 58.26         & 0.447        \\
AudioLDM-2     & 1.64    & 1.86    & 60.32         & 0.432        \\
VATT-LLama     & 2.53    & 3.42    & 75.76         & -            \\
VATT-LLama-T   & 2.07    & 3.25    & 74.89         & 0.376        \\
\hline
\end{tabular}
\end{table*}

\section{Qualitative Examples and Analysis}
\label{sec:qual_samples}
\begin{figure}[tp]
    \centering
    \includegraphics[width=1.0\linewidth]{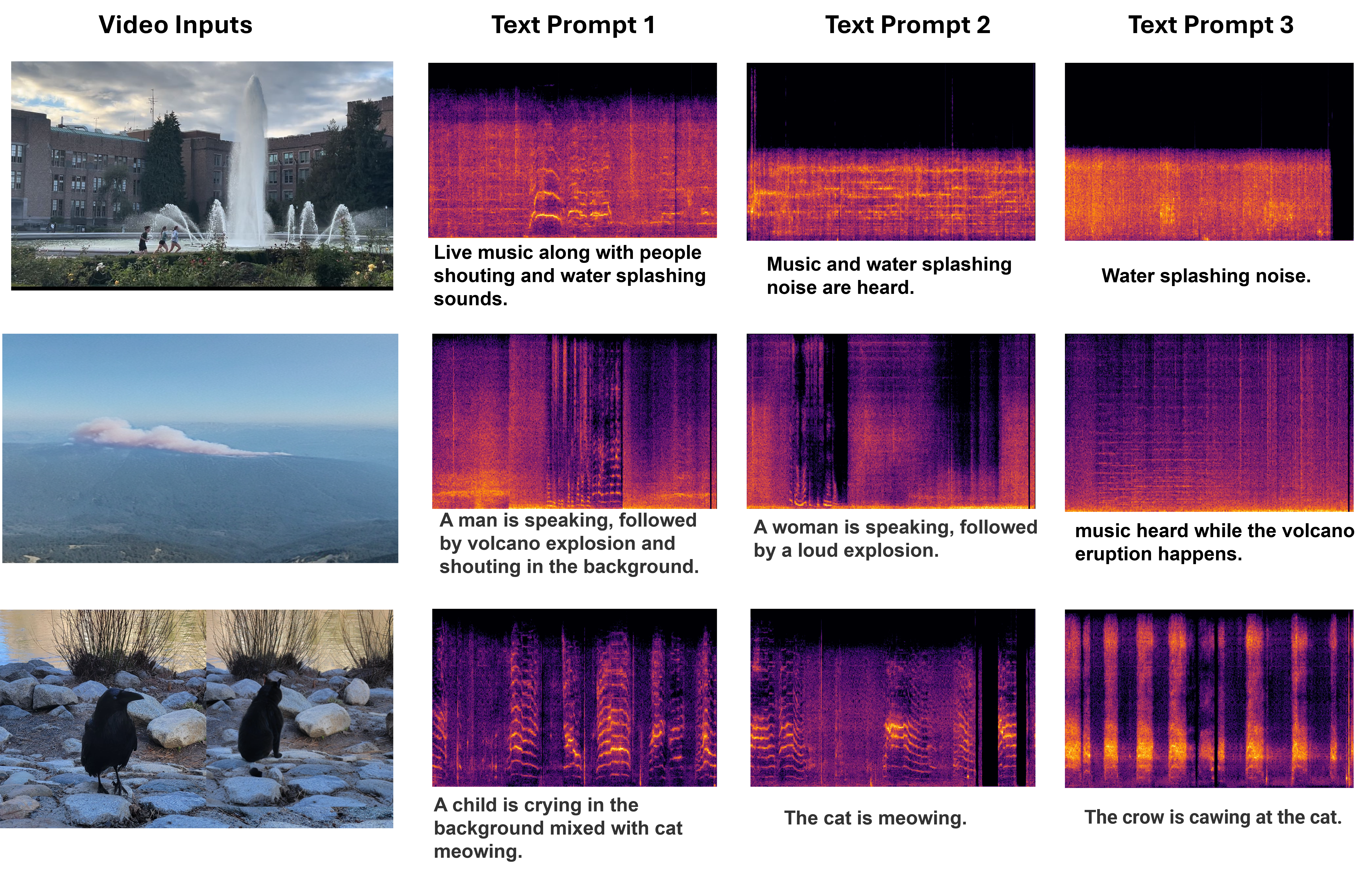}
    \vspace{-2mm}
    \caption{Qualitative samples that showcase text controllability: For same video inputs,  \name is able to generate different sounds that align with the additional text prompts, showcasing its capability of performing controllable generation.}
    \label{fig:text_control}
\end{figure}

\textbf{Controllable audio generation through text prompt:} A unique advantage of our model lies in its capability to control the generated details of audio through text prompts. We show a few samples where different text prompts are applied to the same video to generate different variations of sounds. As shown in Figure \ref{fig:text_control}, our model is able to generate reasonable sounds that are distinct in their semantic meaning but fit both the context of the video and are aligned with the text description. The text prompts shown are all human-written prompts rather than synthetic ones.

\begin{figure}[tp]
    \centering
    \includegraphics[width=1.0\linewidth]{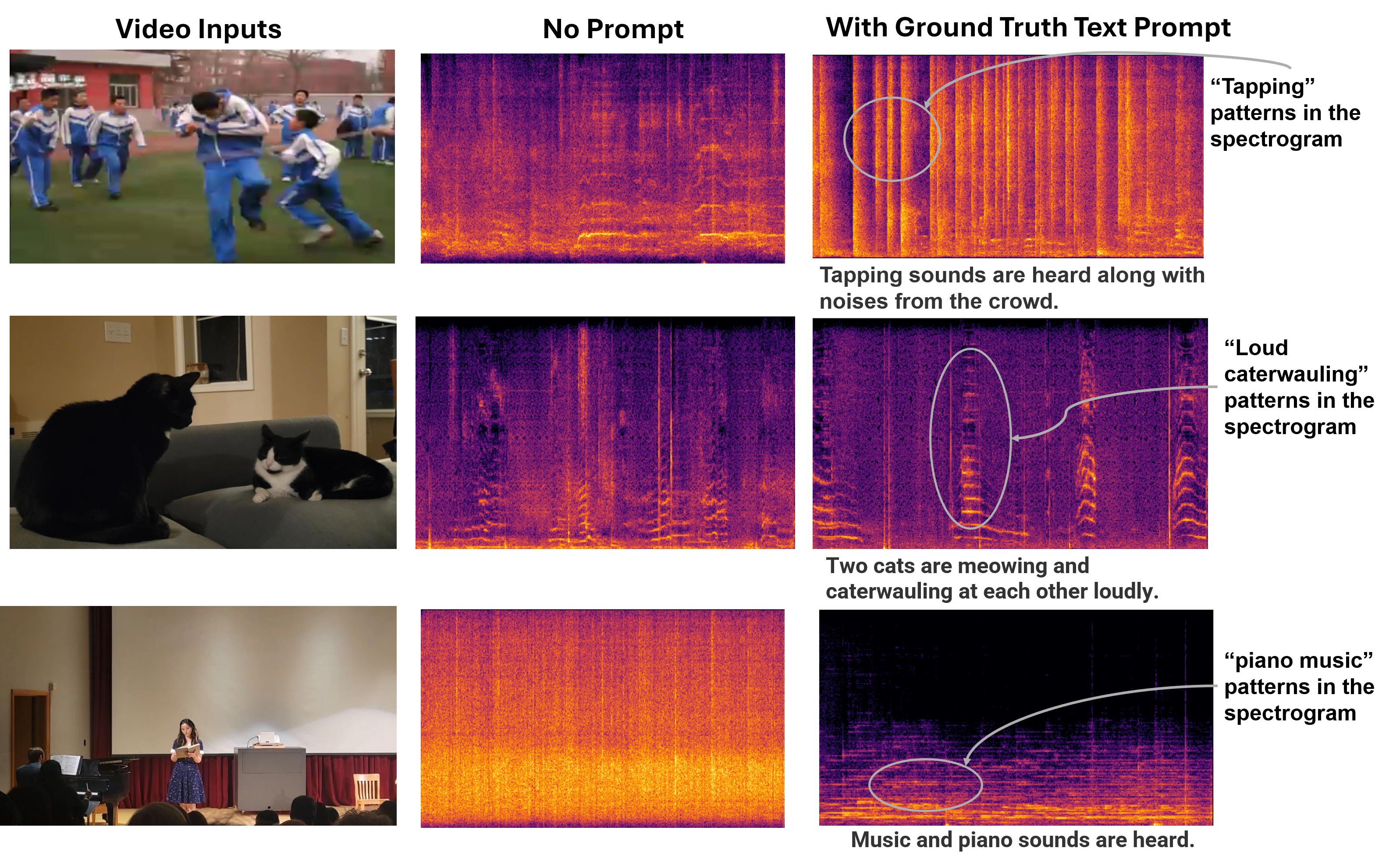}
    \vspace{-2mm}
    \caption{Steering generation towards ground truth audio: For same video inputs, we compare our generation results without text prompt v.s feeding ground truth audio caption as additional prompt.}
    \label{fig:steer_gt}
\end{figure}

\begin{figure}[tp]
    \centering
    \includegraphics[width=1.0\linewidth]{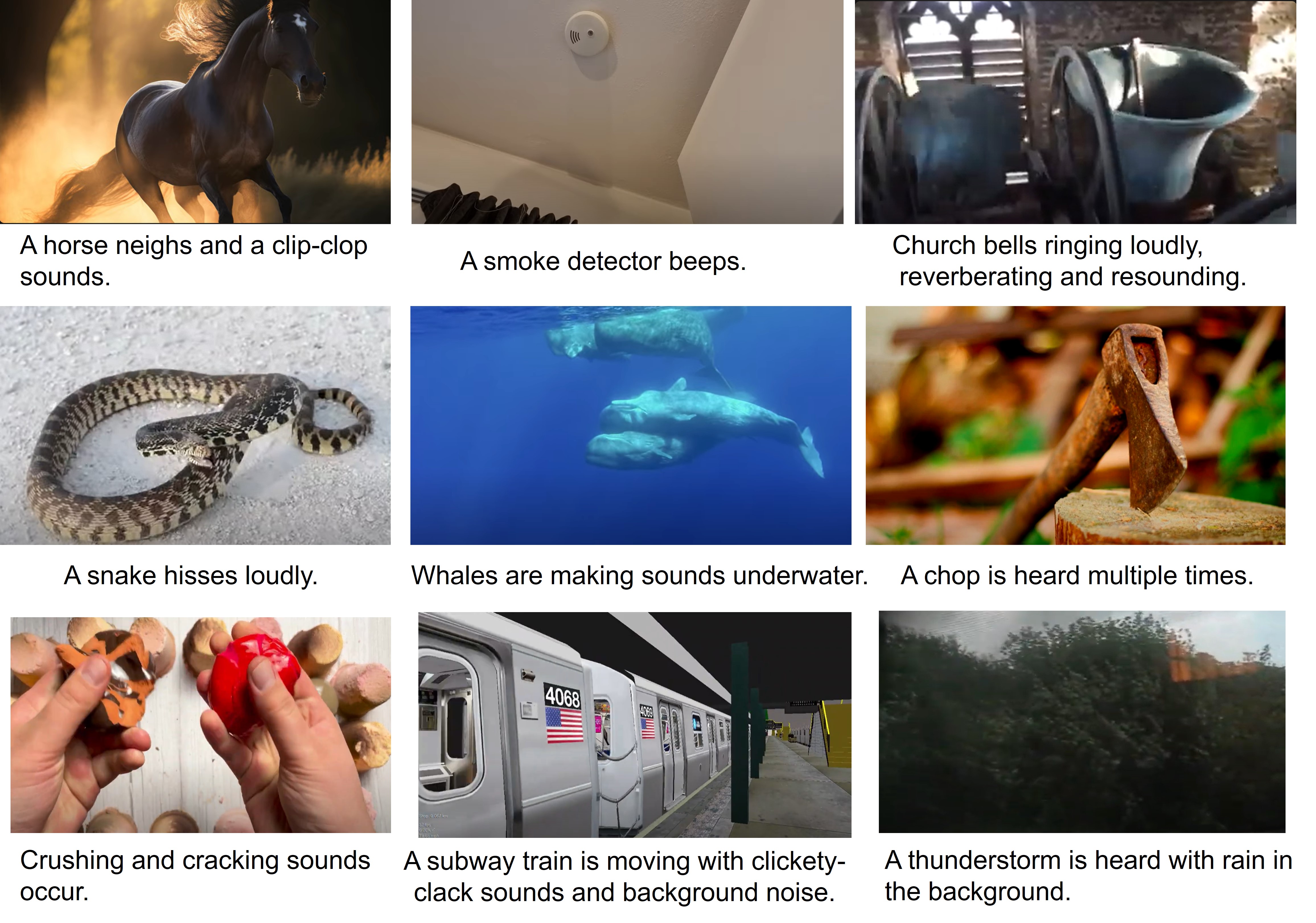}
    \vspace{-2mm} 
    \caption{Video-to-Audio Captioning samples by \name.}
    \label{fig:audio_caption_generation}
\end{figure}

\textbf{\name without prompt v.s \name with ground truth audio captions:} We also compare generation using our model with GT audio captions as prompt versus generation without prompt to understand why the KLD score with prompt outperforms the generation without prompt by a large margin. Upon inspection of generated samples, as shown in Figure \ref{fig:steer_gt}, we find that GT audio captions could steer the model generation towards the GT audio in the test set. For example, the first video shows a man performing with a rope, so the rope tapping sounds (when hitting the ground) should be heard in the video. However, it occurs that the model without prompt fails to capture this important detail, but instead generates noises from the surrounding crowd. Similar cases apply to the other two examples shown in the figure. KLD measures the pairwise difference between generated sounds and GT sounds in the feature space. Therefore, a low score means that the model closely matches the semantic meaning of the GT audio in the original video, which indicates that the model is able to follow the text prompt instruction to generate the desired audio.

\textbf{Video-to-Audio Captioning:} In addition to controllable video-to-audio generation through text, \name is also able to generate audio captions from videos, providing textual suggestions interpreting what sounds could a given video make. As shown in Figure \ref{fig:audio_caption_generation}, \name could produce reasonable audio captions for videos across a variety of audio-visual categories, showcasing the capability of \name Converter in capturing the audio relevant features from the video.

\section{Details and Examples of V2A Instruction Dataset}
\label{sec:v2a_instruction_dataset}
We describe the synthesis procedure of our V2A Instruction dataset. To obtain audio captions for training and evaluating \name, we use existing an existing audio large language model, LTU~\cite{gong2023listen}, which is pretrained on a large-scale audio understanding OpenAQA dataset, including audio from VGGSound and AudioSet-2M. LTU demonstrates strong capability in audio captioning in a zero-shot prompting manner, accurately reflecting what happens in the audio. 

\begin{figure}[tp]
    \centering
    \includegraphics[width=1.1\linewidth]{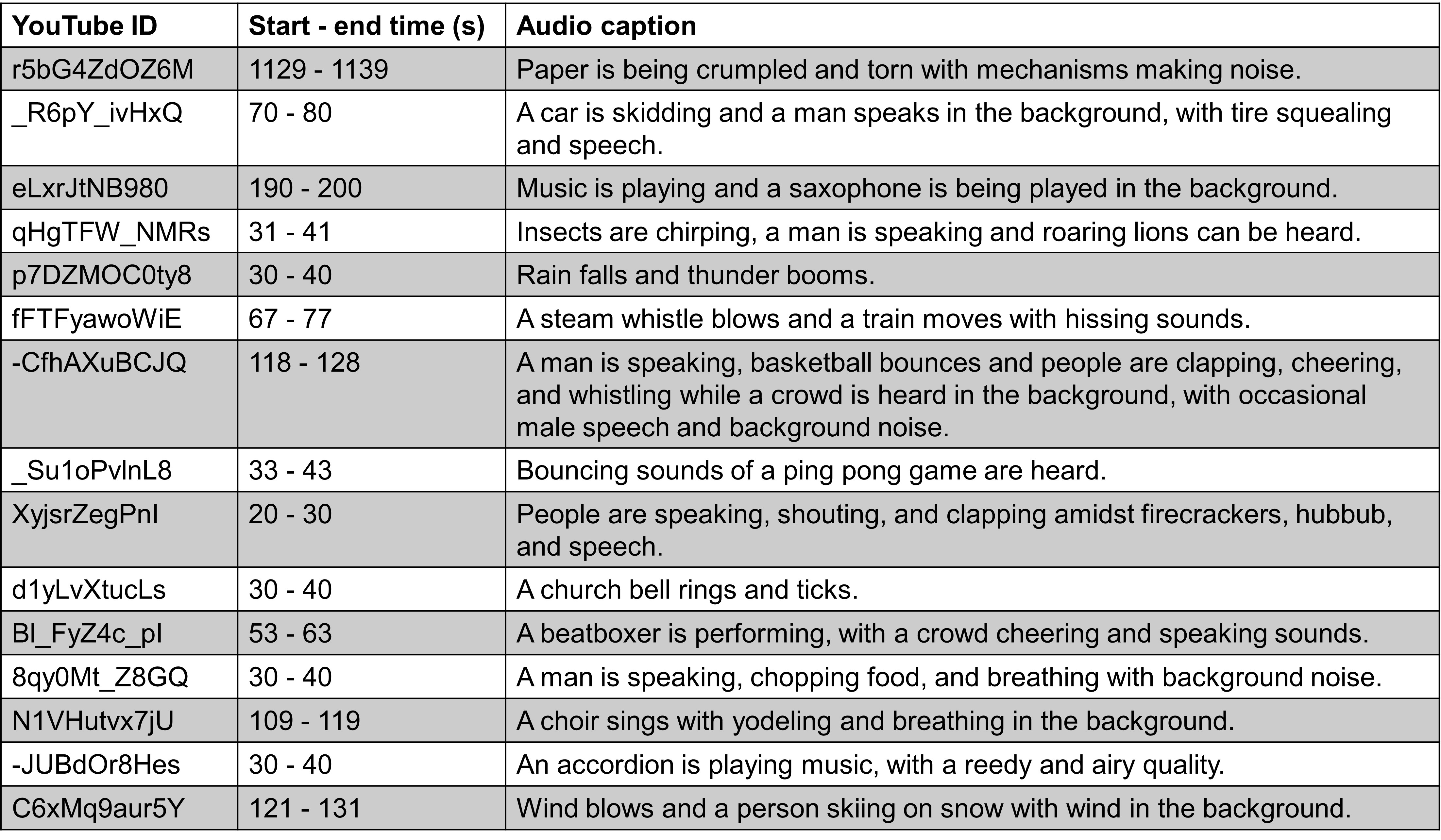}
    \vspace{-2mm}
    \caption{Examples of V2A Instruction Dataset.}
    \label{fig:v2a_instruction}
\end{figure}

Specifically, we adopt the prompt ``Close-ended question: Write an audio caption describing the sound.'' used during LTU training for audio captioning task, and feed 10-second audio from VGGSound and AudioSet dataset as inputs into LTU-13B model (max length 108 version). In figure \ref{fig:v2a_instruction}, we show 15 examples of synthesized captions from videos in VGGSound along with the corresponding video IDs and start-to-end time. The generated captions are clear natural language and faithfully describe the audio content in details, serving as a reliable dataset for training and evaluation.

\section{Additional Implementation Details of \name}
\label{sec:additional_implementation}
\textbf{Data Preprocessing:} For visual inputs, we extract video frames at 5fps rate, resulting in 50 frames for each 10s video. For audio, we extract tokens from audio waveform using pretrained Encodec-16kHz, resulting in a $4 \times 500$ token matrix for each 10s audio. To extract visual features, we resize each video frame to $336 \times 336$ and normalize, and feed into the eva-CLIP-L~\cite{sun2023eva} image encoder. By extracting the mean-pooled vector from the hidden states, we represent each video frame with a 768-dim vector. For textual inputs, we use template text prompts as input to instruct video-to-audio captioning, including 10 human-written prompts. For training the \name Audio, two cases are considered. In unconditional generation case where no additional audio caption is provided, the video along with the one of the template text prompts (shown in Table \ref{tab:text_prompt_template}) are fed as inputs to the model. In conditional generation case where the ground truth audio caption is provided, the caption replaces the template prompt as the textual input to the model. For both cases, the textual inputs are formatted using ``alpaca short'' instruction style, ``\#\#\# Instruction: {instruction}  \#\#\# Response:".

\begin{table*}[h!]
\centering
\begin{tabular}{|c|p{12cm}|}
\hline
\textbf{No.} & \textbf{Text Prompt Template} \\
\hline
1 & Imagine possible sounds for this video. \\
\hline
2 & Describe possible sounds that could match the video. \\
\hline
3 & What audio could be inferred from this video? \\
\hline
4 & What sounds could match the video? \\
\hline
5 & Infer the sounds that match the video. \\
\hline
6 & What audio could best match this video? \\
\hline
7 & What sound events could the video yield? \\
\hline
8 & Caption possible sound events that describe the video. \\
\hline
9 & What sound events make sense to this video? \\
\hline
10 & Imagine audio events that match this video. \\
\hline
\end{tabular}
\caption{Text prompt templates used in video-to-audio captioning and unconditional video-to-audio generation.}
\label{tab:text_prompt_template}
\end{table*}

\textbf{Training:} For \name Converter, we perform V2A Instruction-Tuning with our 1.77M audio captions from VGGSound and Audioset altogether. To project visual features to LLM embedding dimensions, we first apply adaptive pooling along the temporal axis, reducing the temporal dimension from 50 to 10 to conserve GPU memory. Then a linear layer is applied on each time step of the visual features to project 768-dim to the LLM dimension (4096 for LLama and 2048 for Gemma-2B). The training process involves two sub-stages. In the first stage, we turn off the LoRA setting and tune the \name Projector layer only. After 2 epochs, we start the second stage training by tuning both projection weights and LoRA parameters for another 4 epochs. We use AdamW optimizer with a base learning rate of 1e-4, and limit the maximum length of the audio captions to be 108. For both LLama-7B and Gemma-2B, we initialize our model with instruction fine-tuned checkpoints available on huggingface.

For \name Audio, we explore various masking ratio distribution and end up using a truncated Gaussian distribution with a mean of 0.75 and standard deviation of 0.25, truncated between 0.5 and 1.0. To enable classifier-free guidance during sampling, we randomly replace the conditioning features with vectors of all zeros with 10\% rate. The training also takes two sub-stages: i) We first train the model in unconditional generation mode without ground truth audio caption text prompts as inputs until convergence, and then ii) train the model conditioned on GT audio captions as textual inputs (limiting maximum text length to 64). Additionally, we apply two types of data augmentations during the unconditional training phase to facilitate the temporal alignment between audio and video: i) temporal mixup following~\cite{luo2024diff} with a probability of 0.5, ii) temporal rolling on the video features together with audio tokens by same amount of time. We train our model with a base learning rate of 2e-4 with a warmup step of 40k and linear decay schedule with 600k steps. The number of trainable parameters for \name Audio is 415M. We use a batch size of 48 for Gemma model and 36 for LLama model.

All our training procedure are conducted on a single A100 80GB GPU, \name Converter training takes 16 hours and \name Audio takes 3 days in total.

\textbf{Inference:} For Video-to-Audio Captioning, we adopt the generation configuration with temperature of 0.1, top\_p of 0.95, top\_k of 500 and repetition penalty of 1.1 for both LLMs. We use one of the templates ``Describe possible sounds that could match the video.'' to prompt LLMs to generate captions.

For audio generation, we adopt the iterative parallel decoding strategy with total decoding steps of 16, softmax sampling temperate 1.0 with top\_k being 256. For masking sampling with Gumbel top-k strategy, we use initial temperature of 27.5 with linear decay each iteration, where the re-masking ratio follows the cosine schedule, same as~\cite{borsos2023soundstorm, chang2022maskgit}. Following~\cite{mei2023foleygen}, we also use classifier-free guidance~\cite{ho2022classifier} during sampling, and we find a cfg\_scale of 5.0 works best for our model.

\section{Implementation Details of video-to-audio generation baselines}
\label{sec:detail_baselines}
\textbf{SpecVQGAN~\cite{iashin2021taming}:} We follow the open source SpecVQGAN's codebase instructions to extract visual frame-level RGB features at 21.5fps rate using ResNet-50 checkpoints for all VGGSound test videos. Following the evaluation script setup for VGGSound, we generate a 10-second audio waveform per each video in the test set.

\textbf{IM2WAV~\cite{sheffer2023hear}:} Following the open source codebase of IM2WAV, we extract CLIP visual features from videos at 30fps rate for VGGSound test videos. IM2WAV was originally trained to generate 4-second audio. In order to adapt it to generate a 10-second audio, we infer the model with 3 forward passes to obtain three 4-second audio segments without overlap, and concatenate them together and then trim to a 10-second audio.

\textbf{Diff-Foley~\cite{luo2024diff}:} Diff-Foley uses their pretrained CAVP audio-visual model to extract features at 4fps rate. Following their open source codebase, we extract visual features for test videos in VGGSound, and apply their best generation configuration with double guidance scale CFG scale $\omega = 4.5$, CG scale $\gamma = 50$, to generate three non-overlapping 4-second audio segments in the same way as IM2WAV.

\textbf{FoleyGen~\cite{mei2023foleygen}:} The authors do not open source their implementation. We strictly follow the setup in the paper and uses the open source version of Encodec-16kHz to replicate their model. For visual features, we follow their implementation to extract the CLIP features at 1fps rate. FoleyGen is a 24-layers transformer architecture with hidden size of 1024 and 16 heads. Using their best visual attention configuration ``All-frame'' attention along with random  visual condition dropout with a probability of 0.1, we train FoleyGen with specified hyperparameter settings as described in the paper. In the inference stage, we apply classifier-free guidance scale of 3.0 as well as top-k 256 sampling configuration in the paper to generate 10-second video per test video in VGGSound. Upon evaluation, we find that there is a noticeable gap between our implementation results (KLD: 2.89, FAD: 2.59) and reported results in paper (KLD: 2.35, FAD: 1.65). To study where the gap comes from, we use our extracted Encodec tokens to reconstruct the audio waveform in VGGSound test set, and measure the FAD score of reconstructed audio waveform with respect to the ground truth audio. We find that the open source Encodec-16kHz on huggingface could only achieve a FAD score of 1.86, which still falls behind their reported result of 1.65, indicating that the released Encodec model is a sub-optimal version.

\textbf{V2A-Mapper~\cite{wang2024v2a}:} V2A-Mapper is not yet open sourced, but the authors publish their generated audio for 15,446 video samples in VGGSound test set. We therefore download their samples and conduct both objective and subjective experiments based on them.

\section{Details of Evaluation Metrics}
\label{sec:metrics}
\textbf{KLD score}: To compute the KL-Divergence between generated samples and ground truth audio, we adopt the pretrained PaSST~\cite{koutini2021efficient}, an audio transformer classifier, on AudioSet-2M dataset to extract the classification output probabilities. KLD is evaluate in a pairwise manner and we use the mean value over all 15,446 samples as our KLD score.

\textbf{FAD score}:  Fréchet Audio Distance~\cite{kilgour2018fr} (FAD) evaluates the difference between the distributions of generated samples and ground truth samples. Specifically, we adopt the pretrained VGG-ish network on AudioSet-2M dataset to extract the features of generated audio and ground truth audio. Using the multivariate gaussian assumption on extracted features, we compute the mean and covariance of the generation sample set and ground truth audio set, and then apply the Fréchet distance formula to obtain the FAD score. Specifically, to ensure the correctness of computation, we use the original implementation of Google Research's tensorflow version to perform evaluation.

\textbf{Align Acc:} To evaluate temporal alignment and relevance of audio to video, we also incorporate Align Acc metric proposed by~\cite{luo2024diff}. Specifically, Align Acc is computed using a CAVP (contrastive audio-visual pretraining) model by taking video frames along with the audio mel-spectrogram as inputs. The model outputs an accuracy score representing the alignment between audio and video. Following their configuration, we use visual frames at 4 fps rate as visual inputs (40 frames for 10-second video), and convert the 16kHz audio waveform into  mel-spectrogram with FFT Num 1024, mel basis Num 128 and hop size 250, resulting a $640 \times 128$ mel-spectrogram for 10-second audio waveform. 

\textbf{Infer Time:} We benchmark the generation speed on a single A6000 GPU, and measure the average sampling time (in second) for a single 10-second sample. For all baselines and our method, we use a batch size of 1 to run the test over 15,446 test videos on VGGSound. For V2A-Mapper, since we are unable to obtain their source code for testing, we instead report the AudioLDM-L inference speed on a single sample as an approximation of infer Time of V2A-Mapper since the method is a close adaption of AudioLDM.

\textbf{CLAP score} For comparing adherence of generated audio to the text prompt, we use CLAP model to extract the audio and text embeddings, and then measure the cosine similarity between the generated audio embedding and text prompt embedding. For evaluating the video-to-audio captioning, we again apply CLAP to compute the cosine similarity between the generated audio captions and the ground truth audio for the video.

\section{Human Subjective Studies Details} 
\label{sec:human_eval_details}
For human evaluations results on audio generation shown in Table \ref{fig:human_eval}, we used the Amazon Mechanical Turk platform to create a survey and crowdsource responses. We evaluated 100 video samples randomly selected from VGGSound test set. We used stratified sampling such that each video comes from different audio-visual categories. 

To ensure the quality of the survey, we applied constraints on accepted human raters for our survey. In particular, we selected raters that who have historical approval rate of greater than 95\% as well as possess language proficiency in English. Responses that take less than 20 seconds or longer than 10 minutes are excluded from the answers. In addition, no samples could be evaluated twice by the same worker to avoid potential bias.

We set up two types of surveys: audio quality survey and audio-video relevance survey. In audio quality survey, we ask the raters to focus only on the audio quality aspect by providing the question ``In Which video the overall audio quality is better?''. In audio-video relevance survey, a question of ``Which video whose audio is more relevant to and temporally aligned with the video.'' is asked. In both surveys, we ask the rater to choose their preference at a  Likert scale from 1 (strong preference of baseline method) to 5 (strong preference of our method), 5 levels in total.

\begin{figure}[tp]
    \centering
    \includegraphics[width=1.0\linewidth]{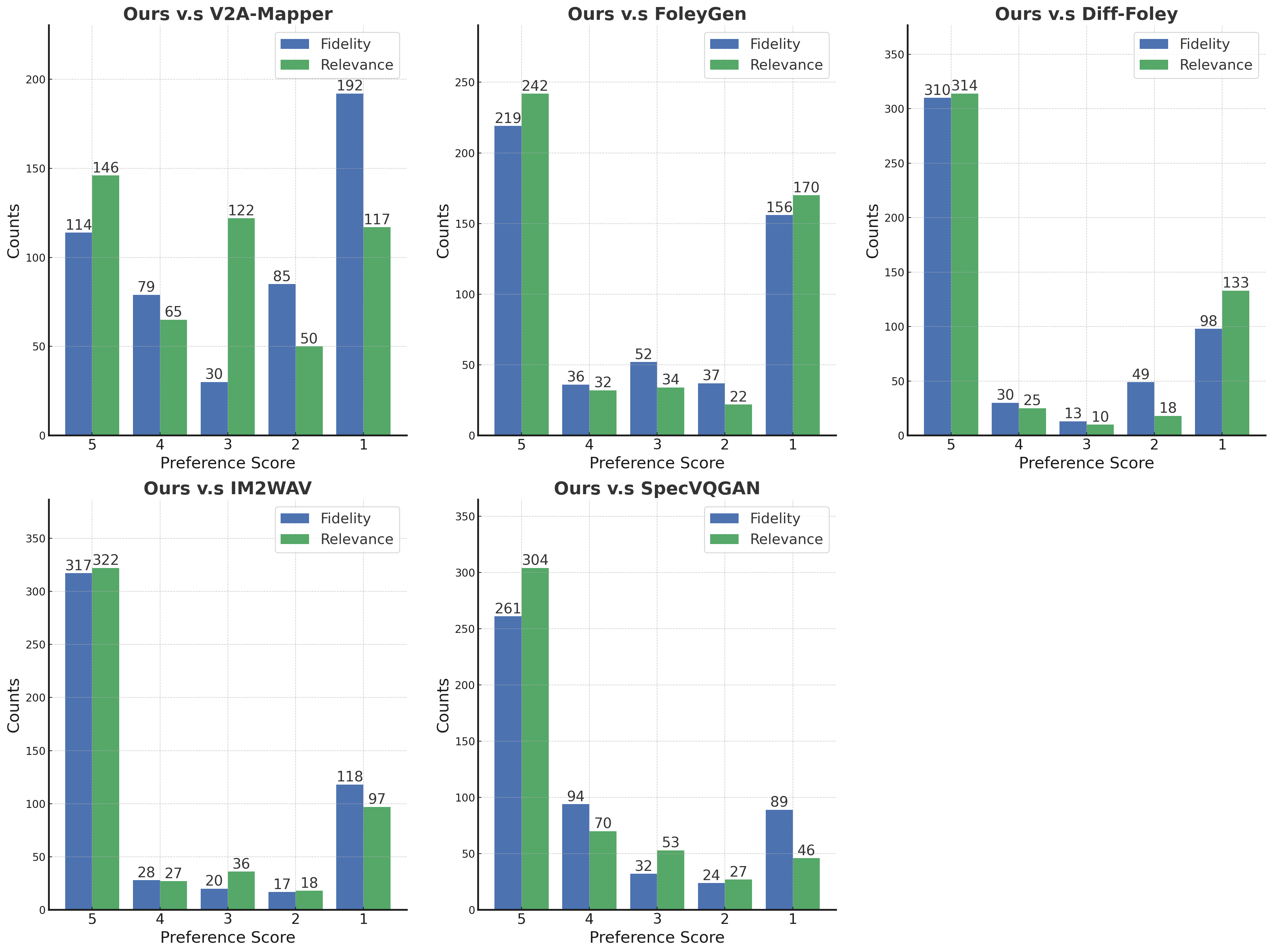}
    \vspace{-2mm}
    \caption{Details of pairwise comparison results between \name and baseline video-to-audio methods.}
    \label{fig:human_eval_stats}
\end{figure}

For each video, we request 5 responses from 5 distinct raters. To comply with the NeurIPS code of conduct and rules of the platform, we pay at a rate of 0.05 USD per each response, satisfying the lowest wage requirement in any region of the world. Upon running the evaluations on all 10 specific surveys (5 methods and each with two evaluation types), we are able to collect 500 valid responses for each pairwise comparison, from 23 distinct raters on average with $21.7 \pm 5.6$ ratings per participant. We show the details of comparison results between our method and baseline methods as bar charts in Figure \ref{fig:human_eval_stats}.

\newpage
\section*{NeurIPS Paper Checklist}

\begin{enumerate}

\item {\bf Claims}
    \item[] Question: Do the main claims made in the abstract and introduction accurately reflect the paper's contributions and scope?
    \item[] Answer: \answerYes{} 
    \item[] Justification: We clearly state the contributions and the experiment results match these contributions.
    \item[] Guidelines:
    \begin{itemize}
        \item The answer NA means that the abstract and introduction do not include the claims made in the paper.
        \item The abstract and/or introduction should clearly state the claims made, including the contributions made in the paper and important assumptions and limitations. A No or NA answer to this question will not be perceived well by the reviewers. 
        \item The claims made should match theoretical and experimental results, and reflect how much the results can be expected to generalize to other settings. 
        \item It is fine to include aspirational goals as motivation as long as it is clear that these goals are not attained by the paper. 
    \end{itemize}

\item {\bf Limitations}
    \item[] Question: Does the paper discuss the limitations of the work performed by the authors?
    \item[] Answer: \answerYes{} 
    \item[] Justification: We mention the limitations and the room for improvements in the Conclusion section.
    \item[] Guidelines:
    \begin{itemize}
        \item The answer NA means that the paper has no limitation while the answer No means that the paper has limitations, but those are not discussed in the paper. 
        \item The authors are encouraged to create a separate "Limitations" section in their paper.
        \item The paper should point out any strong assumptions and how robust the results are to violations of these assumptions (e.g., independence assumptions, noiseless settings, model well-specification, asymptotic approximations only holding locally). The authors should reflect on how these assumptions might be violated in practice and what the implications would be.
        \item The authors should reflect on the scope of the claims made, e.g., if the approach was only tested on a few datasets or with a few runs. In general, empirical results often depend on implicit assumptions, which should be articulated.
        \item The authors should reflect on the factors that influence the performance of the approach. For example, a facial recognition algorithm may perform poorly when image resolution is low or images are taken in low lighting. Or a speech-to-text system might not be used reliably to provide closed captions for online lectures because it fails to handle technical jargon.
        \item The authors should discuss the computational efficiency of the proposed algorithms and how they scale with dataset size.
        \item If applicable, the authors should discuss possible limitations of their approach to address problems of privacy and fairness.
        \item While the authors might fear that complete honesty about limitations might be used by reviewers as grounds for rejection, a worse outcome might be that reviewers discover limitations that aren't acknowledged in the paper. The authors should use their best judgment and recognize that individual actions in favor of transparency play an important role in developing norms that preserve the integrity of the community. Reviewers will be specifically instructed to not penalize honesty concerning limitations.
    \end{itemize}

\item {\bf Theory Assumptions and Proofs}
    \item[] Question: For each theoretical result, does the paper provide the full set of assumptions and a complete (and correct) proof?
    \item[] Answer: \answerNo{} 
    \item[] Justification: Our paper is mainly application and experiment-driven.
    \item[] Guidelines:
    \begin{itemize}
        \item The answer NA means that the paper does not include theoretical results. 
        \item All the theorems, formulas, and proofs in the paper should be numbered and cross-referenced.
        \item All assumptions should be clearly stated or referenced in the statement of any theorems.
        \item The proofs can either appear in the main paper or the supplemental material, but if they appear in the supplemental material, the authors are encouraged to provide a short proof sketch to provide intuition. 
        \item Inversely, any informal proof provided in the core of the paper should be complemented by formal proofs provided in appendix or supplemental material.
        \item Theorems and Lemmas that the proof relies upon should be properly referenced. 
    \end{itemize}

    \item {\bf Experimental Result Reproducibility}
    \item[] Question: Does the paper fully disclose all the information needed to reproduce the main experimental results of the paper to the extent that it affects the main claims and/or conclusions of the paper (regardless of whether the code and data are provided or not)?
    \item[] Answer: \answerYes{} 
    \item[] Justification: We include sufficient details of implementations of our models, baseline models, and also evaluation metrics along with human evaluation surveys.
    \item[] Guidelines:
    \begin{itemize}
        \item The answer NA means that the paper does not include experiments.
        \item If the paper includes experiments, a No answer to this question will not be perceived well by the reviewers: Making the paper reproducible is important, regardless of whether the code and data are provided or not.
        \item If the contribution is a dataset and/or model, the authors should describe the steps taken to make their results reproducible or verifiable. 
        \item Depending on the contribution, reproducibility can be accomplished in various ways. For example, if the contribution is a novel architecture, describing the architecture fully might suffice, or if the contribution is a specific model and empirical evaluation, it may be necessary to either make it possible for others to replicate the model with the same dataset, or provide access to the model. In general. releasing code and data is often one good way to accomplish this, but reproducibility can also be provided via detailed instructions for how to replicate the results, access to a hosted model (e.g., in the case of a large language model), releasing of a model checkpoint, or other means that are appropriate to the research performed.
        \item While NeurIPS does not require releasing code, the conference does require all submissions to provide some reasonable avenue for reproducibility, which may depend on the nature of the contribution. For example
        \begin{enumerate}
            \item If the contribution is primarily a new algorithm, the paper should make it clear how to reproduce that algorithm.
            \item If the contribution is primarily a new model architecture, the paper should describe the architecture clearly and fully.
            \item If the contribution is a new model (e.g., a large language model), then there should either be a way to access this model for reproducing the results or a way to reproduce the model (e.g., with an open-source dataset or instructions for how to construct the dataset).
            \item We recognize that reproducibility may be tricky in some cases, in which case authors are welcome to describe the particular way they provide for reproducibility. In the case of closed-source models, it may be that access to the model is limited in some way (e.g., to registered users), but it should be possible for other researchers to have some path to reproducing or verifying the results.
        \end{enumerate}
    \end{itemize}

\item {\bf Open access to data and code}
    \item[] Question: Does the paper provide open access to the data and code, with sufficient instructions to faithfully reproduce the main experimental results, as described in supplemental material?
    \item[] Answer: \answerYes{} 
    \item[] Justification: 
    Code and synthetic data that we generated for training models can be obtained for limited use upon request from the corresponding author. We also aim to release the code and synthetic data that we generated for training models as public repository on the Github upon obtaining involved approvals.
    \item[] Guidelines:
    \begin{itemize}
        \item The answer NA means that paper does not include experiments requiring code.
        \item Please see the NeurIPS code and data submission guidelines (\url{https://nips.cc/public/guides/CodeSubmissionPolicy}) for more details.
        \item While we encourage the release of code and data, we understand that this might not be possible, so “No” is an acceptable answer. Papers cannot be rejected simply for not including code, unless this is central to the contribution (e.g., for a new open-source benchmark).
        \item The instructions should contain the exact command and environment needed to run to reproduce the results. See the NeurIPS code and data submission guidelines (\url{https://nips.cc/public/guides/CodeSubmissionPolicy}) for more details.
        \item The authors should provide instructions on data access and preparation, including how to access the raw data, preprocessed data, intermediate data, and generated data, etc.
        \item The authors should provide scripts to reproduce all experimental results for the new proposed method and baselines. If only a subset of experiments are reproducible, they should state which ones are omitted from the script and why.
        \item At submission time, to preserve anonymity, the authors should release anonymized versions (if applicable).
        \item Providing as much information as possible in supplemental material (appended to the paper) is recommended, but including URLs to data and code is permitted.
    \end{itemize}

\item {\bf Experimental Setting/Details}
    \item[] Question: Does the paper specify all the training and test details (e.g., data splits, hyperparameters, how they were chosen, type of optimizer, etc.) necessary to understand the results?
    \item[] Answer: \answerYes{} 
    \item[] Justification: We specify all details of training, inference, dataset and evaluation metrics necessary to achieve the results we claim in the experiments. 
    \item[] Guidelines:
    \begin{itemize}
        \item The answer NA means that the paper does not include experiments.
        \item The experimental setting should be presented in the core of the paper to a level of detail that is necessary to appreciate the results and make sense of them.
        \item The full details can be provided either with the code, in appendix, or as supplemental material.
    \end{itemize}

\item {\bf Experiment Statistical Significance}
    \item[] Question: Does the paper report error bars suitably and correctly defined or other appropriate information about the statistical significance of the experiments?
    \item[] Answer: \answerNA{} 
    \item[] Justification: Our models are evaluated on large-scale test set and the variations should be small and negligible.
    \item[] Guidelines:
    \begin{itemize}
        \item The answer NA means that the paper does not include experiments.
        \item The authors should answer "Yes" if the results are accompanied by error bars, confidence intervals, or statistical significance tests, at least for the experiments that support the main claims of the paper.
        \item The factors of variability that the error bars are capturing should be clearly stated (for example, train/test split, initialization, random drawing of some parameter, or overall run with given experimental conditions).
        \item The method for calculating the error bars should be explained (closed form formula, call to a library function, bootstrap, etc.)
        \item The assumptions made should be given (e.g., Normally distributed errors).
        \item It should be clear whether the error bar is the standard deviation or the standard error of the mean.
        \item It is OK to report 1-sigma error bars, but one should state it. The authors should preferably report a 2-sigma error bar than state that they have a 96\% CI, if the hypothesis of Normality of errors is not verified.
        \item For asymmetric distributions, the authors should be careful not to show in tables or figures symmetric error bars that would yield results that are out of range (e.g. negative error rates).
        \item If error bars are reported in tables or plots, The authors should explain in the text how they were calculated and reference the corresponding figures or tables in the text.
    \end{itemize}

\item {\bf Experiments Compute Resources}
    \item[] Question: For each experiment, does the paper provide sufficient information on the computer resources (type of compute workers, memory, time of execution) needed to reproduce the experiments?
    \item[] Answer: \answerYes{} 
    \item[] Justification: We include the details of computing devices, memory and time of execution used to conduct training and evaluation of our method.
    \item[] Guidelines:
    \begin{itemize}
        \item The answer NA means that the paper does not include experiments.
        \item The paper should indicate the type of compute workers CPU or GPU, internal cluster, or cloud provider, including relevant memory and storage.
        \item The paper should provide the amount of compute required for each of the individual experimental runs as well as estimate the total compute. 
        \item The paper should disclose whether the full research project required more compute than the experiments reported in the paper (e.g., preliminary or failed experiments that didn't make it into the paper). 
    \end{itemize}
    
\item {\bf Code Of Ethics}
    \item[] Question: Does the research conducted in the paper conform, in every respect, with the NeurIPS Code of Ethics \url{https://neurips.cc/public/EthicsGuidelines}?
    \item[] Answer: \answerYes{} 
    \item[] Justification: We have reviewed and carefully followed the NeurIPS Code of Ethics.
    \item[] Guidelines:
    \begin{itemize}
        \item The answer NA means that the authors have not reviewed the NeurIPS Code of Ethics.
        \item If the authors answer No, they should explain the special circumstances that require a deviation from the Code of Ethics.
        \item The authors should make sure to preserve anonymity (e.g., if there is a special consideration due to laws or regulations in their jurisdiction).
    \end{itemize}

\item {\bf Broader Impacts}
    \item[] Question: Does the paper discuss both potential positive societal impacts and negative societal impacts of the work performed?
    \item[] Answer: \answerYes{} 
    \item[] Justification: We discuss broader impacts in Appendix \ref{sec:broader_impacts}.
    \item[] Guidelines:
    \begin{itemize}
        \item The answer NA means that there is no societal impact of the work performed.
        \item If the authors answer NA or No, they should explain why their work has no societal impact or why the paper does not address societal impact.
        \item Examples of negative societal impacts include potential malicious or unintended uses (e.g., disinformation, generating fake profiles, surveillance), fairness considerations (e.g., deployment of technologies that could make decisions that unfairly impact specific groups), privacy considerations, and security considerations.
        \item The conference expects that many papers will be foundational research and not tied to particular applications, let alone deployments. However, if there is a direct path to any negative applications, the authors should point it out. For example, it is legitimate to point out that an improvement in the quality of generative models could be used to generate deepfakes for disinformation. On the other hand, it is not needed to point out that a generic algorithm for optimizing neural networks could enable people to train models that generate Deepfakes faster.
        \item The authors should consider possible harms that could arise when the technology is being used as intended and functioning correctly, harms that could arise when the technology is being used as intended but gives incorrect results, and harms following from (intentional or unintentional) misuse of the technology.
        \item If there are negative societal impacts, the authors could also discuss possible mitigation strategies (e.g., gated release of models, providing defenses in addition to attacks, mechanisms for monitoring misuse, mechanisms to monitor how a system learns from feedback over time, improving the efficiency and accessibility of ML).
    \end{itemize}
    
\item {\bf Safeguards}
    \item[] Question: Does the paper describe safeguards that have been put in place for responsible release of data or models that have a high risk for misuse (e.g., pretrained language models, image generators, or scraped datasets)?
    \item[] Answer: \answerYes{} 
    \item[] Justification: We discuss the need for safeguards regarding the negative impact associated with multi-modal generative modeling and propose precautions and guidelines to prevent misuse. These details are included as part of the Broader Impact section in Appendix \ref{sec:broader_impacts}.
    \item[] Guidelines:
    \begin{itemize}
        \item The answer NA means that the paper poses no such risks.
        \item Released models that have a high risk for misuse or dual-use should be released with necessary safeguards to allow for controlled use of the model, for example by requiring that users adhere to usage guidelines or restrictions to access the model or implementing safety filters. 
        \item Datasets that have been scraped from the Internet could pose safety risks. The authors should describe how they avoided releasing unsafe images.
        \item We recognize that providing effective safeguards is challenging, and many papers do not require this, but we encourage authors to take this into account and make a best faith effort.
    \end{itemize}

\item {\bf Licenses for existing assets}
    \item[] Question: Are the creators or original owners of assets (e.g., code, data, models), used in the paper, properly credited and are the license and terms of use explicitly mentioned and properly respected?
    \item[] Answer: \answerYes{} 
    \item[] Justification: We cited all open-source works that we use including the data, models and codes.
    \item[] Guidelines:
    \begin{itemize}
        \item The answer NA means that the paper does not use existing assets.
        \item The authors should cite the original paper that produced the code package or dataset.
        \item The authors should state which version of the asset is used and, if possible, include a URL.
        \item The name of the license (e.g., CC-BY 4.0) should be included for each asset.
        \item For scraped data from a particular source (e.g., website), the copyright and terms of service of that source should be provided.
        \item If assets are released, the license, copyright information, and terms of use in the package should be provided. For popular datasets, \url{paperswithcode.com/datasets} has curated licenses for some datasets. Their licensing guide can help determine the license of a dataset.
        \item For existing datasets that are re-packaged, both the original license and the license of the derived asset (if it has changed) should be provided.
        \item If this information is not available online, the authors are encouraged to reach out to the asset's creators.
    \end{itemize}

\item {\bf New Assets}
    \item[] Question: Are new assets introduced in the paper well documented and is the documentation provided alongside the assets?
    \item[] Answer: \answerYes{} 
    \item[] Justification: We incorporated a section in supplementary text describing our synthetic data and placed samples. We aim to release the synthetic data that we generated for training upon obtaining involved approvals.
    \item[] Guidelines:
    \begin{itemize}
        \item The answer NA means that the paper does not release new assets.
        \item Researchers should communicate the details of the dataset/code/model as part of their submissions via structured templates. This includes details about training, license, limitations, etc. 
        \item The paper should discuss whether and how consent was obtained from people whose asset is used.
        \item At submission time, remember to anonymize your assets (if applicable). You can either create an anonymized URL or include an anonymized zip file.
    \end{itemize}

\item {\bf Crowdsourcing and Research with Human Subjects}
    \item[] Question: For crowdsourcing experiments and research with human subjects, does the paper include the full text of instructions given to participants and screenshots, if applicable, as well as details about compensation (if any)? 
    \item[] Answer: \answerYes{} 
    \item[] Justification: We included detailed description of human study instructions, and and the setup of the of the study.
    \item[] Guidelines:
    \begin{itemize}
        \item The answer NA means that the paper does not involve crowdsourcing nor research with human subjects.
        \item Including this information in the supplemental material is fine, but if the main contribution of the paper involves human subjects, then as much detail as possible should be included in the main paper. 
        \item According to the NeurIPS Code of Ethics, workers involved in data collection, curation, or other labor should be paid at least the minimum wage in the country of the data collector. 
    \end{itemize}

\item {\bf Institutional Review Board (IRB) Approvals or Equivalent for Research with Human Subjects}
    \item[] Question: Does the paper describe potential risks incurred by study participants, whether such risks were disclosed to the subjects, and whether Institutional Review Board (IRB) approvals (or an equivalent approval/review based on the requirements of your country or institution) were obtained?
    \item[] Answer: \answerYes{} 
    \item[] Justification: We submitted IRB approval request to our institution for review and it was approved.

    \item[] Guidelines:
    \begin{itemize}
        \item The answer NA means that the paper does not involve crowdsourcing nor research with human subjects.
        \item Depending on the country in which research is conducted, IRB approval (or equivalent) may be required for any human subjects research. If you obtained IRB approval, you should clearly state this in the paper. 
        \item We recognize that the procedures for this may vary significantly between institutions and locations, and we expect authors to adhere to the NeurIPS Code of Ethics and the guidelines for their institution. 
        \item For initial submissions, do not include any information that would break anonymity (if applicable), such as the institution conducting the review.
    \end{itemize}

\end{enumerate}

\end{document}